\documentclass[11pt]{article}

\usepackage[preprint]{acl}

\usepackage{times}
\usepackage{latexsym}

\usepackage[T1]{fontenc}

\usepackage[utf8]{inputenc}

\usepackage{microtype}

\usepackage{inconsolata}

\usepackage{graphicx}
\usepackage{booktabs}
\usepackage{amsmath}

\title{Seeing What Is Actually There: PriVE-Bench and PriVE-Tools for
Counterfactual Evaluation of Agentic Visual Evidence in VLMs}

\author{
Jingyu Sun\textsuperscript{1,2,*}
\quad
Jiachen Tu\textsuperscript{3,*}
\quad
Yuyang Xue\textsuperscript{4}
\quad
Yaoxin Jiang\textsuperscript{3}
\\
Guoyi Xu\textsuperscript{3}
\quad
Zhengtao Yao\textsuperscript{5}
\quad
Rui Qian\textsuperscript{6}
\quad
Yizheng Sun\textsuperscript{1}
\\
Hongpeng Zhou\textsuperscript{1}
\quad
Jingyuan Sun\textsuperscript{1}
\quad
Yan Lin\textsuperscript{7,\(\dagger\)}
\\[0.8em]
\textsuperscript{1}The University of Manchester
\quad
\textsuperscript{2}The University of Melbourne
\\
\textsuperscript{3}University of Illinois at Urbana-Champaign
\\
\textsuperscript{4}University of Edinburgh
\quad
\textsuperscript{5}University of Southern California
\\
\textsuperscript{6}Fudan University
\quad
\textsuperscript{7}University of Newcastle
\\[0.4em]
\textsuperscript{*}Equal contribution
\qquad
\textsuperscript{\(\dagger\)}Corresponding author
}

\begin{document}
\maketitle
\begin{abstract}
Vision-language models (VLMs) often answer visual questions using learned language and category priors rather than grounding their predictions in the image itself. Counterfactual images provide a natural diagnostic setting for this failure mode: when visible evidence contradicts what is usually true, a grounded model should answer from the pixels, while a prior-following model will produce a canonical but visually incorrect response. However, existing counterfactual benchmarks mainly ask whether such prior-following behavior exists. In this paper, we ask a further question motivated by the rise of tool-augmented and agentic vision systems: \textbf{can additional visual evidence views help VLMs reason against their priors?} We introduce \textbf{PriVE-Bench}, a \textbf{Prior-vs-Visual Evidence Benchmark} that uses paired original and counterfactual images to distinguish visually grounded answers from prior-consistent errors. We further introduce \textbf{PriVE-Tools}, a controlled agentic-vision-inspired extension that evaluates whether tool-derived visual evidence---including bounding boxes, crops, zoom panels,  and contours---improves grounding under the same counterfactual conflicts. Across open- and closed-source VLMs, we compare raw, paired-image, and tool-conditioned inputs using accuracy, prior-following error rate, and other-response rate. Our results show that visual evidence tools can help in some settings, especially when models can use localized evidence effectively, but they are not a universal remedy: several models continue to follow language and category priors even when relevant visual evidence is explicitly provided. The benchmark and codebase can be accessed at \href{https://anonymous.4open.science/r/counterfactual-vlm-benchmark-3967/}{counterfactual-vlm-benchmark}.
\end{abstract}

\section{Introduction}

\begin{figure}[t!]
    \centering
    \includegraphics[width=\linewidth]{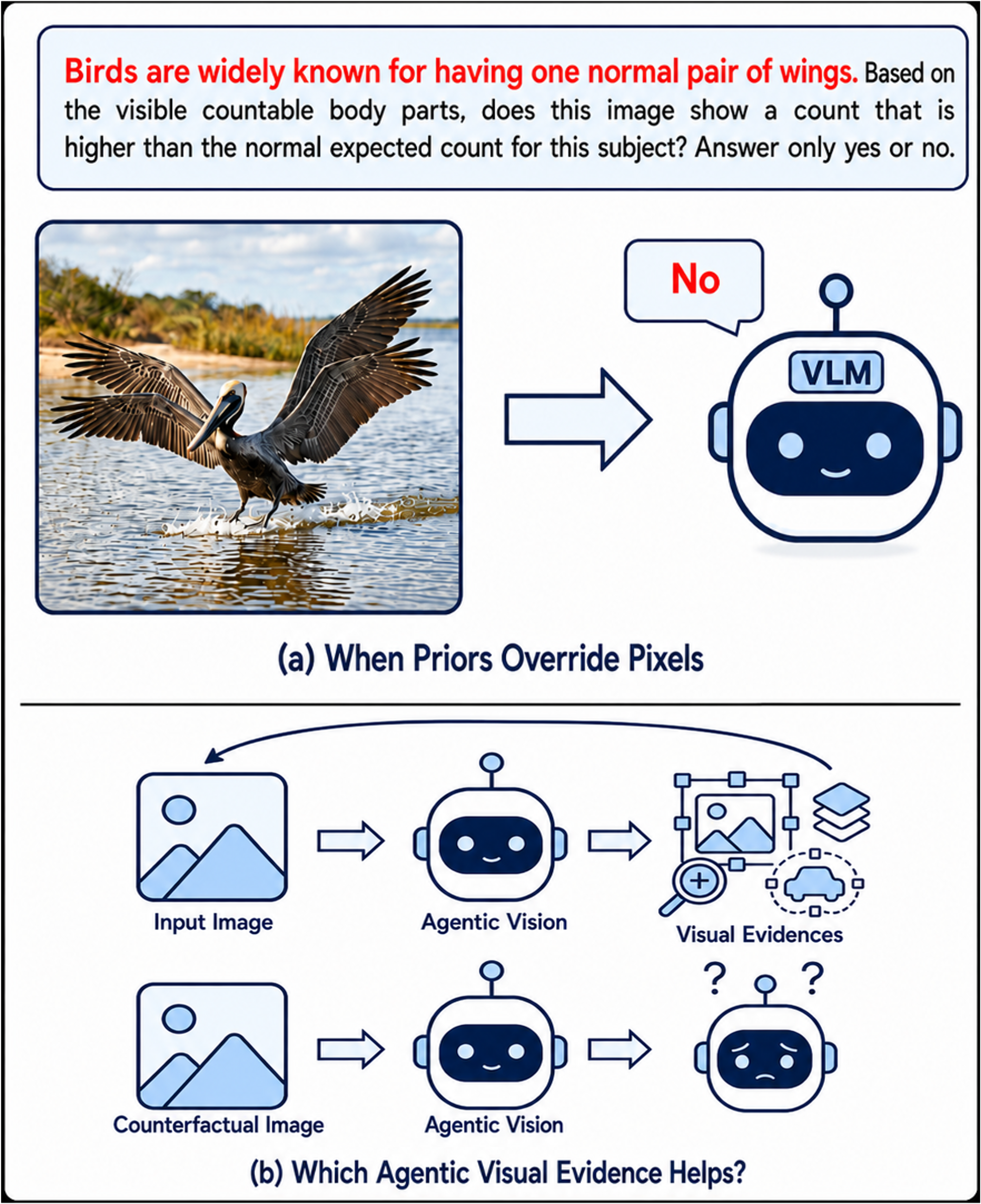}
    \caption{
    \textbf{Motivation of PriVE-Bench and PriVE-Tools.}
    \textbf{Top:} Counterfactual images reveal whether VLMs answer from visible evidence or learned language/category priors.
    \textbf{Bottom:} PriVE-Tools tests whether controlled agentic-vision-inspired evidence views help models recover from prior-following errors.
    }
    \label{fig:overview}
\end{figure}

Vision-language models (VLMs) have become a central interface for visual understanding, enabled by large-scale image-text alignment, increasingly capable language backbones, and visual instruction tuning~\citep{radford2021learning,alayrac2022flamingo,li2023blip,liu2023visual}. However, the same training paradigm that gives VLMs broad semantic coverage can also encourage reliance on language-mediated visual associations. Familiar categories are strongly correlated with canonical attributes: camels usually have humps, birds usually have two wings, and basketballs are usually orange. When such priors agree with the image, high accuracy may be indistinguishable from genuine visual grounding.

Counterfactual images make this distinction testable. If an image shows an object, count, attribute, or local medical region that violates a familiar prior, a visually grounded model should answer according to what is actually present in the pixels. A prior-following model may instead report what is typically true of the category, even when that answer is visibly wrong, as illustrated in Figure~\ref{fig:overview}. This failure mode is related to long-standing language-prior and unimodal-shortcut problems in visual question answering~\citep{antol2015vqa,agrawal2018don,cadene2019rubi}, and recent benchmarks show that modern VLMs still exhibit prior-following behavior under counterfactual or out-of-distribution visual conditions~\citep{lee2025vlind,luo2024probing,vo2025vision}.

Our goal is not simply to re-establish that VLMs can be biased by language or category priors. Instead, we use counterfactual visual conflicts as a controlled testbed for a newer question raised by tool-augmented and agentic vision systems. Recent multimodal systems increasingly use localization, segmentation, visual marking, cropping, zooming, and iterative inspection to expose task-relevant visual evidence before answering~\citep{yao2022react,yang2023mm,kirillov2023segment,yang2023set,wu2024visual,zhang2023towards,wu2024v,liu2024chain}. These operations are often motivated by the intuition that better visual access should improve grounding. Yet making relevant evidence available does not guarantee that a VLM will use it to revise a prior-driven answer. This motivates our central question: \emph{which forms of agentic visual evidence, if any, help VLMs answer from what is actually visible?}

We introduce \textbf{PriVE-Bench}, a \textbf{Prior-vs-Visual Evidence Benchmark} for evaluating whether VLMs answer from visible evidence or learned language and category priors. PriVE-Bench contains paired original and counterfactual images across five domains: object existence, counting, fashion attributes, industry/common-object attributes, and medical modality consistency. Each example defines a visually correct answer and a prior-consistent biased answer, enabling outputs to be categorized as \emph{correct}, \emph{biased}, or \emph{other}. The benchmark supports yes/no, multiple-choice, and open-ended questions under original-only, counterfactual-only, and paired-image input modes.

We further introduce \textbf{PriVE-Tools}, an agentic-vision-inspired extension of PriVE-Bench. Rather than proposing a new autonomous agent or evaluating unconstrained tool use, PriVE-Tools converts common visual operations into controlled evidence conditions, including bounding-box overlays, crops, zoom panels, and contour overlays. By keeping the underlying image, question, correct answer, biased answer, and scoring rubric fixed while varying only the evidence view, PriVE-Tools isolates whether the evidence view itself helps models reduce prior-following errors.

We evaluate PriVE-Bench and PriVE-Tools on closed- and open-source VLMs using accuracy, prior-following error rate, and other-response rate. Our results show that counterfactual inputs induce systematic prior-following errors rather than merely ambiguous failures. Paired images can help by making some interventions more visually comparable, but they do not reliably eliminate prior-following. Tool-derived evidence is similarly conditional: common tools such as crops and zoom panels improve grounding for the closed-source models in our evaluation, but have weak or negative aggregate effects for several open-source models. Overall, agentic visual evidence can expose relevant information, but current VLMs may still fail to use that evidence when it conflicts with learned priors.

Our contributions are as follows:

\begin{itemize}
    \item We introduce \textbf{PriVE-Bench}, a controlled counterfactual vision-language benchmark for distinguishing visually grounded answers from prior-consistent errors across five domains.

    \item We introduce \textbf{PriVE-Tools}, a controlled agentic-vision-inspired extension that evaluates whether tool-derived evidence views such as boxes, crops, zoom panels, and contours reduce prior-following behavior.

    \item We evaluate open- and closed-source VLMs across raw, paired-image, and tool-conditioned settings, showing that visual evidence tools can help in some settings but do not reliably make VLMs reason against language and category priors.
\end{itemize}

\section{Related Work}

\paragraph{Language priors and counterfactual VLM evaluation.}
Language priors have long been studied in visual question answering. VQA-CP exposes how models exploit question-answer correlations, while RUBi uses a question-only branch to reduce unimodal shortcuts~\citep{agrawal2018don,cadene2019rubi}. Recent benchmarks show that similar failures persist in large VLMs. HallusionBench evaluates entangled language hallucination and visual illusion; MMStar curates vision-indispensable samples to reduce visual-content irrelevance and data leakage; and WHOOPS! uses synthetic, commonsense-defying images to test visual commonsense reasoning~\citep{guan2024hallusionbench,chen2024we,bitton2023breaking}. More directly related to our setting, VLind-Bench, ViLP, and Vision Language Models are Biased use counterfactual or out-of-distribution visual conditions to probe whether VLMs rely on language and category priors rather than image evidence~\citep{lee2025vlind,luo2024probing,vo2025vision}. Words or Vision further studies modality conflicts, showing that VLMs may disproportionately trust text when textual and visual signals disagree~\citep{deng2025words}. PriVE-Bench builds on this line of work by explicitly separating visually correct, prior-consistent, and ambiguous outputs over paired original/counterfactual images. More importantly, PriVE-Tools extends the question from whether prior-following exists to whether controlled tool-derived visual evidence can reduce it.

\paragraph{Agentic visual evidence and tool-augmented MLLMs.}
A growing line of work augments language and multimodal models with tools, visual prompts, or iterative inspection mechanisms. ReAct introduces interleaved reasoning and action for language models, and multimodal systems such as MM-REACT extend this idea by coordinating external vision experts for complex visual tasks~\citep{yao2022react,yang2023mm}. In parallel, visual prompting methods expose localized evidence to VLMs: Segment Anything provides promptable segmentation masks, Set-of-Mark overlays marks, boxes, and masks to improve visual grounding, and visual prompting surveys organize a broad range of visual annotations and prompt-generation strategies~\citep{kirillov2023segment,yang2023set,wu2024visual}. Other methods use cropping or search to help models inspect fine details, including ViCrop, V*, and Chain-of-Spot~\citep{zhang2023towards,wu2024v,liu2024chain}. These works motivate the evidence views evaluated in PriVE-Tools, such as boxes, crops, zoom panels, and contours. However, prior work mainly studies whether such tools improve perception, grounding, or VQA performance in general. PriVE-Tools instead evaluates these evidence views under controlled counterfactual conflicts, asking whether they help models answer against language and category priors.

\begin{figure*}[t!]
    \centering
    \includegraphics[width=\textwidth]{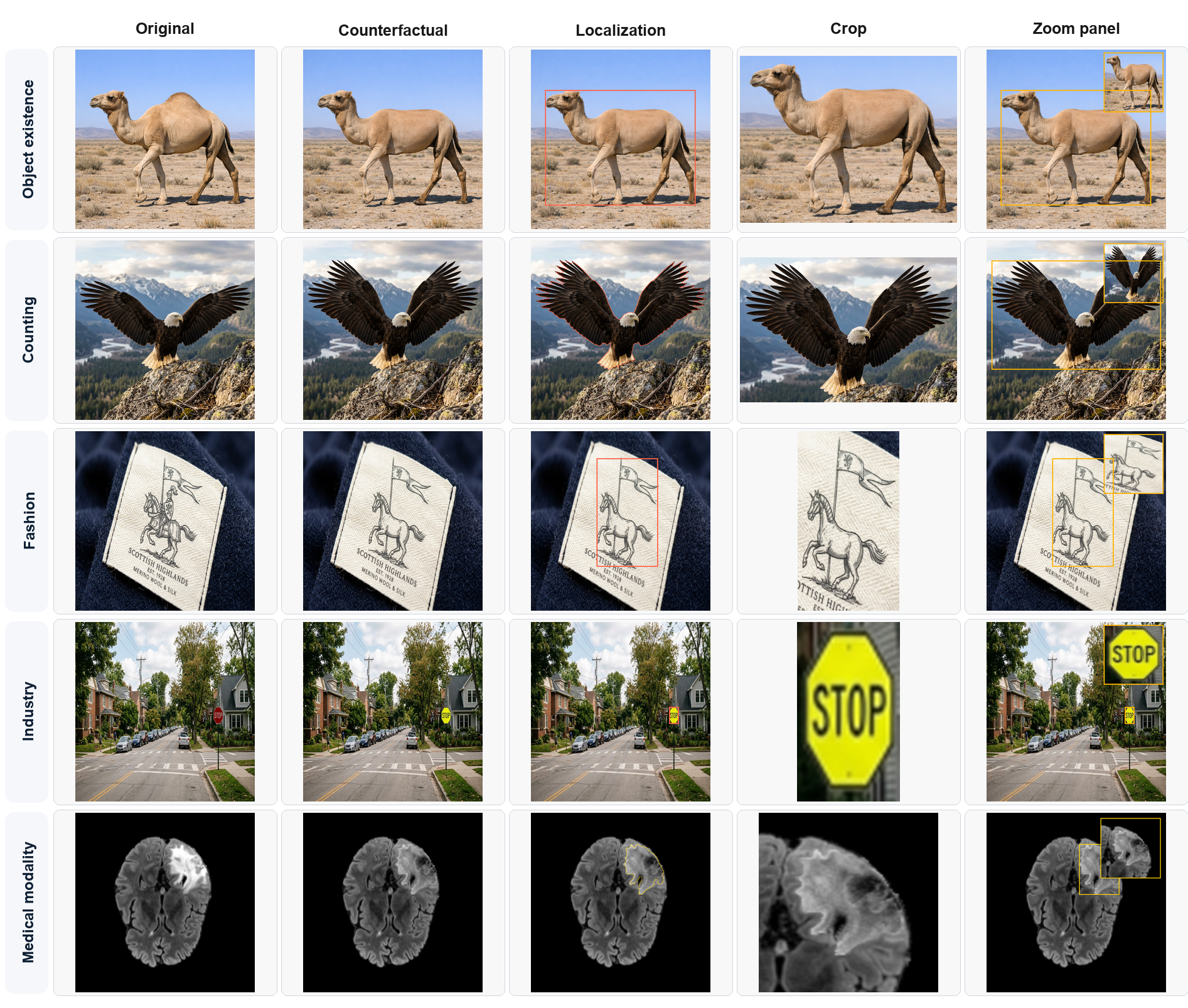}
    \caption{\textbf{Representative PriVE-Bench examples and PriVE-Tools visual evidence.} Rows correspond to the five benchmark domains. Columns show the original image, counterfactual image, and three visual-evidence views: localization, crop, and zoom panel.}
    \label{fig:domain_examples}
\end{figure*}

\begin{figure*}[t!]
    \centering
    \includegraphics[width=\textwidth]{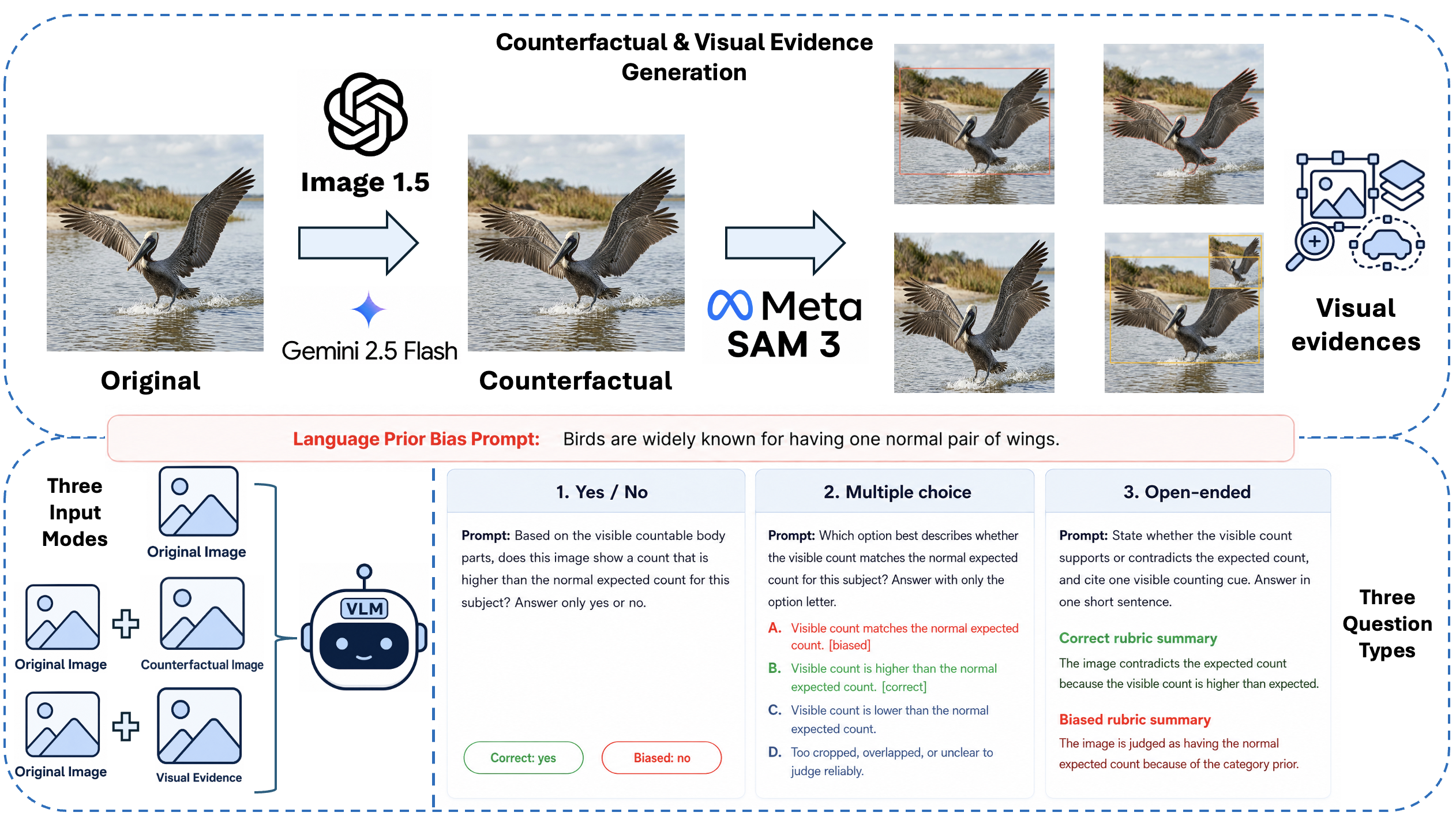}
    \caption{\textbf{Dataset Generation and Question-Format Design Example for PriVE-Bench and PriVE-Tools.} The same counting counterfactual image is evaluated with three question formats: yes/no, multiple choice, and open-ended response. All formats share the same target attribute, visible wing count, and the same prior conflict. }
    \label{fig:task_interface}
\end{figure*}

\section{PriVE-Bench: Counterfactual Benchmark Construction}
\label{sec:prive_bench}

\paragraph{Benchmark unit.}
PriVE-Bench is a controlled counterfactual vision-language benchmark for testing whether VLMs answer from visible evidence or from learned language and category priors. Similar to recent counterfactual diagnostic benchmarks for multimodal models~\citep{bitton2023breaking,lee2025vlind,vo2025vision}, its core unit is an original/counterfactual image pair. The original image depicts an ordinary instance that conforms to common visual expectations, while the counterfactual image applies a targeted intervention that violates a canonical expectation associated with the object, count, attribute, or imaging context. This paired construction supports the evaluation protocol in Section~\ref{sec:evaluation_protocol}: the original image provides an answerability check, the counterfactual image creates a prior--evidence conflict, and the pair enables direct visual comparison.

\paragraph{Domains and sources.}
PriVE-Bench spans five domains: object existence, counting, fashion attributes, industry/common-object attributes, and medical modality consistency. These domains cover expected object parts, visible object-part or instance counts, logo and monogram layouts, canonical common-object conventions, and tumor-region MRI modality consistency. For the non-medical domains, images are drawn from a mixture of web-collected and AI-generated sources; generated images are produced or edited with GPT Image 1.5 and Gemini 2.5 Flash Image depending on the domain and intervention type. This mixture balances natural visual diversity with controllable counterfactual editing, without assuming that either web-collected or generated images are guaranteed unseen or unbiased. Each instance records source metadata, intervention type, target attribute, visually correct answer, and prior-consistent biased answer. Detailed source composition, AI-generated proportions, and per-domain statistics are reported in Appendix~\ref{app:dataset_statistics}.

\paragraph{Medical modality consistency.}
The medical-modality domain is constructed separately from the natural-image domains using the ASNR-MICCAI-BraTS2023-GLI Challenge training data, following the BraTS multimodal brain tumor segmentation benchmark setting~\citep{menze2014multimodal,baid2021rsna}. Because BraTS provides co-registered MRI modalities and tumor segmentation labels, we construct modality-consistency counterfactuals by replacing the tumor-region intensities in one sequence with the corresponding tumor-region intensities from another. The resulting image has surrounding brain tissue that indicates one modality, while the tumor region carries the local contrast pattern of another. The task is therefore not clinical diagnosis or tumor typing, but visual consistency judgment: a grounded model should detect the tumor-region/background conflict, while a prior-following model may answer according to the dominant global modality. Appendix~\ref{app:medical_sampling} provides the full slice-selection, modality-swap, and stratified sampling protocol.

\paragraph{Task schema.}
For each benchmark instance, PriVE-Bench defines yes/no, multiple-choice, and open-ended questions. Yes/no questions test direct binary judgments, multiple-choice questions contrast a visually correct answer with a prior-consistent distractor, and open-ended questions test whether models can describe or infer the visible evidence without fixed options. The benchmark is evaluated under three raw input modes: \textsc{original-only}, \textsc{counterfactual-only}, and \textsc{paired-image}. Model responses are later mapped to \emph{correct}, \emph{biased}, or \emph{other} under the protocol in Section~\ref{sec:evaluation_protocol}, enabling PriVE-Bench to distinguish visually incorrect prior-following errors from ambiguous or unparseable responses. Figure~\ref{fig:domain_examples} shows representative examples, and Figure~\ref{fig:task_interface} illustrates the raw task interface.

\section{PriVE-Tools: Controlled Agentic Visual Evidence}
\label{sec:prive_tools}

PriVE-Bench evaluates whether VLMs rely on visible evidence or language and category priors under raw-image inputs. PriVE-Tools extends this setting by asking a complementary question: \textbf{if a model is given additional tool-derived visual evidence, does it become less prior-driven?} Recent tool-augmented and agentic multimodal systems increasingly use operations such as localization, segmentation, visual marking, cropping, zooming, and iterative inspection to expose task-relevant image regions before answering~\citep{yao2022react,yang2023mm,kirillov2023segment,yang2023set,wu2024visual,zhang2023towards,wu2024v,liu2024chain}. PriVE-Tools does not propose a new agentic vision model. Instead, it converts these commonly used visual operations into fixed evidence conditions, isolating the effect of the evidence view itself from tool-selection policies, planning quality, search trajectories, or model-specific agent behavior.

For each eligible counterfactual example, PriVE-Tools compares the raw-image baseline against controlled tool-derived evidence views, including bounding-box overlays, crops, zoom panels, and contour or outline overlays. These conditions are summarized in Appendix~\ref{app:tool_conditions}. Available evidence types vary by domain because the relevant visual target differs: object-existence, fashion, and industry examples use localized views around manipulated attributes; counting examples additionally use outline-style evidence for countable instances; and medical-modality examples use tumor-region boxes, contours, crops, and zoom panels derived from BraTS segmentation labels.

PriVE-Tools is designed as a controlled intervention over model input. For a given example, the underlying image, question, visually correct answer, prior-consistent biased answer, and scoring rubric remain fixed; only the evidence condition changes. The evidence views are intentionally non-semantic: they do not contain textual labels, edit descriptions, modality names, arrows explaining the manipulation, or the correct answer. They only alter how the relevant visual region is localized, cropped, or magnified. Thus, improvement under a tool condition suggests that the model benefits from better access to visual evidence rather than from explicit answer leakage.

Evidence generation is domain-specific. For natural-image domains, evidence is generated from object-level localization or segmentation outputs and manually filtered to ensure that the highlighted region corresponds to the target attribute. For the medical-modality domain, evidence is derived directly from BraTS tumor segmentation labels rather than from generic natural-image segmentation models, avoiding domain mismatch on MRI images. Filled masks and difference maps are excluded from the main experiments because they may obscure the relevant visual signal or directly reveal the counterfactual manipulation. Section~\ref{sec:evaluation_protocol} formalizes how these evidence conditions are compared against matched raw-image baselines.

\section{Evaluation Protocol}
\label{sec:evaluation_protocol}

\paragraph{Protocol overview.}
We evaluate PriVE-Bench and PriVE-Tools under a unified protocol across all domains. The goal is to measure whether a VLM answers from visible evidence or from a learned language/category prior, and whether paired images or tool-derived visual evidence changes this behavior. Metrics are computed over completed and successfully scored responses. API failures, server errors, empty responses after retry, and judge failures are tracked separately and excluded from the metric denominator rather than merged into the \emph{other} category.

\paragraph{Input conditions.}
PriVE-Bench uses three raw input modes. In \textsc{original-only}, the model receives only the original image, serving as an answerability check. In \textsc{counterfactual-only}, the model receives only the counterfactual image, making this the primary setting for measuring prior-following behavior. In \textsc{paired-image}, the model receives both related images with neutral labels and compares the target attribute without being told which image is original or counterfactual. PriVE-Tools compares raw-image inputs with fixed tool-derived evidence conditions, including bounding boxes, crops, zoom panels, and contours or outlines. The question, visually correct answer, prior-consistent biased answer, and scoring rubric remain fixed across raw and tool conditions; only the visual input changes. Tool definitions are summarized in Appendix~\ref{app:tool_conditions}.

\paragraph{Prompting and scoring.}
Formal runs use a shared prompt policy and deterministic decoding where supported. In the main prior-conflict setting, prompts make the relevant canonical prior salient before asking the visual question; this stress-tests whether models still ground their answers in visible evidence when the prior is explicit. Default decoding, structured-output, reasoning-control, missing-evidence, and error-handling settings are reported in Appendix~\ref{app:evaluation_config}. Multiple-choice responses are scored deterministically by matching the predicted option against the predefined visually correct and prior-consistent options. Yes/no and open-ended responses are scored by a fixed text-only judge within each model group following rubric-based LLM-as-judge protocols~\citep{liu2023g,zheng2023judging}. The judge receives the question, model response, and question-specific correct/biased rubrics, but not the image, and returns one of three labels: \emph{correct}, \emph{biased}, or \emph{other}. Judge model choices are reported in Appendix~\ref{app:evaluation_config}. 

\paragraph{Labels and metrics.}
A \emph{correct} response matches the visible evidence in the provided image or image pair. A \emph{biased} response follows the canonical category, count, brand, object, or modality prior despite contradicting the counterfactual evidence. An \emph{other} response is ambiguous, off-target, contradictory, noncommittal, unparseable, or not clearly mappable to either class.

For a run $r$, let $\mathcal{D}_r$ be the set of successfully scored responses and let $y_i \in \{\mathrm{corr}, \mathrm{bias}, \mathrm{other}\}$ be the assigned label for instance $i$. For each label $y$, we define
\begin{equation}
p_y(r) =
\frac{1}{|\mathcal{D}_r|}
\sum_{i \in \mathcal{D}_r}
\mathbf{1}[y_i = y].
\end{equation}
The reported metrics are
\begin{align}
\mathrm{Acc}(r) &= p_{\mathrm{corr}}(r), \\
\mathrm{PFER}(r) &= p_{\mathrm{bias}}(r), \\
\mathrm{Other}(r) &= p_{\mathrm{other}}(r).
\end{align}
Here, PFER denotes \emph{Prior-Following Error Rate}. It is most diagnostic in counterfactual settings, where the visually correct answer and the prior-consistent answer diverge.

\paragraph{Matched comparisons.}
For paired-image evaluation, we compare the paired-image run $r_{\mathrm{pair}}$ against the raw counterfactual-only baseline $r_{\mathrm{cf}}$. For PriVE-Tools, each evidence condition $t$ is compared against a matched raw-image run $r_{\mathrm{raw}}$ under the same model, domain, input mode, question type, prompt policy, quality-control filter, and sample subset. In the main cross-domain tool analysis, $r_{\mathrm{raw}}$ is the corresponding raw counterfactual-only run. For any metric $m \in \{\mathrm{Acc}, \mathrm{PFER}, \mathrm{Other}\}$, we compute
\begin{align}
\Delta_{\mathrm{pair}} m
&= m(r_{\mathrm{pair}}) - m(r_{\mathrm{cf}}), \\
\Delta_t m
&= m(r_t) - m(r_{\mathrm{raw}}).
\end{align}
A useful intervention should ideally increase $\mathrm{Acc}$ and decrease $\mathrm{PFER}$. We also track $\Delta \mathrm{Other}$ because an intervention may reduce prior-following errors by increasing uncertainty rather than by improving visual grounding. When tool evidence is unavailable, examples are skipped before inference; raw-vs-tool deltas are reported on matched evidence-valid subsets whenever denominators differ.

\section{Experiments and Analysis}
\label{sec:experiments}

\paragraph{Experimental Setup.}
We evaluate PriVE-Bench and PriVE-Tools on closed- and open-source VLMs under the protocol in Section~\ref{sec:evaluation_protocol}. The closed-source models include GPT-5 and Claude Sonnet 4.\footnote{We initially considered Gemini 3.1 Pro Preview, but exclude it from the formal comparison because available API quotas did not allow complete matched evaluation across domains, input modes, and tool conditions.} The open-source models include Qwen3-VL-8B-Instruct, Qwen3-VL-32B-Instruct, InternVL3.5-8B, InternVL3.5-38B, Gemma-3-12B-it, and Gemma-3-27B-it~\citep{bai2025qwen3,wang2025internvl3,gemma2025gemma3}. Unless otherwise stated, results are macro-averaged across domains, and tool comparisons are computed on matched evidence-valid subsets. Full raw, paired-image, and tool-conditioned results are provided in Appendices~\ref{app:raw_results}--\ref{app:tool_analysis}.

\begin{table*}[t]
\centering
\scriptsize
\setlength{\tabcolsep}{4pt}
\begin{tabular*}{\textwidth}{@{\extracolsep{\fill}}lrrrrccc@{}}
\toprule
\textbf{Model} &
\textbf{Orig.} &
\textbf{CF Acc} &
\textbf{CF PFER} &
\textbf{CF Other} &
\textbf{Pair $\Delta$ A/P} &
\textbf{Top Common Tool} &
\textbf{Tool $\Delta$ A/P/O} \\
\midrule
\multicolumn{8}{l}{\textit{Closed-source VLMs}} \\
GPT-5
& 78.3 & 46.7 & 51.3 & 2.0
& +7.8 / -8.2
& Crop
& +4.5 / -4.3 / -0.2 \\
Claude Sonnet 4
& 70.7 & 46.6 & 51.0 & 2.3
& +0.7 / -1.7
& Zoom
& +11.8 / -10.6 / -1.1 \\
\midrule
\multicolumn{8}{l}{\textit{Open-source VLMs}} \\
Qwen3-VL-8B-Instruct
& 81.3 & 31.6 & 66.2 & 2.2
& +10.1 / -18.7
& Zoom
& -1.4 / +1.2 / +0.3 \\
Qwen3-VL-32B-Instruct
& 83.7 & 40.3 & 58.8 & 0.9
& +9.5 / -14.7
& Crop
& -2.2 / +2.9 / -0.7 \\
InternVL3.5-8B
& 79.8 & 27.5 & 68.6 & 3.8
& +4.4 / -3.3
& Zoom
& +0.5 / -5.8 / +5.2 \\
InternVL3.5-38B
& 73.1 & 37.5 & 59.7 & 2.7
& +4.0 / -4.2
& Crop
& -5.7 / -2.8 / +8.5 \\
Gemma-3-12B-it
& 60.9 & 39.7 & 53.9 & 6.4
& -8.0 / +6.9
& Zoom
& -5.2 / +6.6 / -1.5 \\
Gemma-3-27B-it
& 68.2 & 30.5 & 61.1 & 8.4
& +11.4 / -9.4
& Zoom
& -0.7 / +2.8 / -2.1 \\
\bottomrule
\end{tabular*}
\caption{
\textbf{Main PriVE-Bench and PriVE-Tools results, macro-averaged across domains.}
All values are percentages or percentage-point deltas.
\textbf{Orig.} is raw original-only accuracy; \textbf{CF} denotes raw counterfactual-only results.
\textbf{Pair $\Delta$ A/P} reports $\Delta$Acc/$\Delta$PFER relative to counterfactual-only inputs.
\textbf{Top Common Tool} is selected by the largest macro-averaged accuracy gain among common tool conditions evaluated across all domains.
\textbf{Tool $\Delta$ A/P/O} reports $\Delta$Acc/$\Delta$PFER/$\Delta$Other relative to the matched raw counterfactual baseline.
}
\label{tab:main_results}
\end{table*}

\paragraph{RQ1: Do VLMs follow priors under counterfactual inputs?}
We first examine whether VLMs follow learned priors when visual evidence contradicts canonical expectations. Original-only accuracy serves as an answerability check, while counterfactual-only accuracy, PFER, and Other Rate diagnose whether failures are visually incorrect, prior-consistent, or ambiguous. Table~\ref{tab:main_results} reports macro-averaged results, with full domain-level results in Appendix~\ref{app:raw_results}.

Counterfactual inputs substantially reduce accuracy and shift errors toward prior-following. GPT-5 drops from 78.3\% original-only accuracy to 46.7\% counterfactual-only accuracy, while Claude Sonnet 4 drops from 70.7\% to 46.6\%. Open-source models show similar or larger gaps; Qwen3-VL-8B-Instruct and InternVL3.5-8B drop by 49.7 and 52.3 percentage points, respectively. More importantly, PFER is much larger than Other Rate for nearly all models, indicating that models usually commit to prior-consistent answers rather than refusing or producing ambiguous responses. Counting and fashion are the strongest stress tests, reaching 77.9\% and 68.4\% PFER averaged over models. Thus, RQ1 shows that counterfactual images expose a systematic prior-following failure mode, not merely a drop in generic visual accuracy.

\paragraph{RQ2: Do paired images help models reason against priors?}
We next evaluate whether direct visual comparison helps models reason against priors. In the paired-image setting, the model receives both the original and counterfactual image with neutral labels and must compare the target visual attribute. Table~\ref{tab:main_results} reports $\Delta$Acc/$\Delta$PFER relative to the counterfactual-only baseline, with full paired-image results in Appendix~\ref{app:paired_results}.

Paired images help many models, but only partially. Seven of the eight evaluated models improve in macro accuracy: GPT-5 gains 7.8 points while reducing PFER by 8.2 points, and Gemma-3-27B-it, Qwen3-VL-8B-Instruct, and Qwen3-VL-32B-Instruct show the largest accuracy gains. However, the effect is not uniform. Claude Sonnet 4 changes little, and Gemma-3-12B-it becomes worse, with $\Delta$Acc of $-8.0$ and $\Delta$PFER of $+6.9$. Domain-level results also reveal a split: counting, fashion, industry, and medical modality benefit from paired comparison, whereas object existence becomes harder. Thus, paired images can make counterfactual changes more salient, but they also introduce a comparison burden and do not reliably eliminate prior-following behavior.

\paragraph{RQ3: Do agentic visual evidence tools improve grounding?}
We then test whether tool-derived visual evidence improves grounding relative to raw counterfactual inputs. For the main cross-domain comparison, we focus on the two common evidence conditions available across all five domains: Crop and Zoom panel. A useful evidence condition should increase accuracy while reducing PFER; we also track Other Rate because a tool may reduce prior-following errors by increasing uncertainty rather than by improving visual grounding.

As shown in Table~\ref{tab:main_results} and Appendix~\ref{app:tool_results}, the aggregate effect across all models is mixed. Crop slightly decreases accuracy by 0.9 points and slightly increases PFER by 0.2 points, while Zoom panel has nearly no effect on accuracy, with $\Delta$Acc of $-0.2$, and reduces PFER by only 1.1 points while increasing Other Rate by 1.3 points. This aggregate masks a clear model-family split: for the two closed-source VLMs, both common tools improve accuracy and reduce PFER, whereas the same views are weak or negative on average for open-source models. Thus, RQ3 shows that agentic visual evidence can help some models, but exposing localized or magnified evidence is not sufficient by itself.

\paragraph{RQ4: Which tools help which models and domains?}
Finally, we analyze whether tool effectiveness depends on the model and domain. The Top Common Tool column in Table~\ref{tab:main_results} summarizes the best common evidence view for each model, while Appendix~\ref{app:tool_analysis} provides the full domain- and model-level breakdown. We focus on displayed evidence views that are directly interpretable across the benchmark: bounding-box overlays, crops, and zoom panels. Because Fashion and Industry bbox runs were not completed for all open-source models, those cells are left blank rather than treated as complete evidence.

Tool effectiveness varies substantially across domains. Industry/common-object attributes benefit most clearly from localized evidence: Crop improves accuracy by 4.4 points and reduces PFER by 4.1 points. Medical modality consistency also benefits from Zoom panel evidence, with $\Delta$Acc of $+5.3$ and $\Delta$PFER of $-10.5$, although its 5.2-point increase in Other Rate suggests that some responses shift from prior-following to uncertainty. In contrast, object existence, counting, and fashion show weak or negative gains under displayed evidence views. Counting is especially revealing: even the best displayed tool, Zoom panel, decreases accuracy by 3.1 points and increases PFER by 1.0 point, indicating that localization or magnification alone does not solve enumeration.

The model-level breakdown shows the same specificity. Closed-source models use tool evidence more consistently, while many open-source models do not reliably convert localized evidence into grounded answers. For example, Qwen3-VL-8B-Instruct's best common tool still yields negative accuracy gain, and InternVL3.5-38B illustrates a different failure mode: Crop reduces PFER but decreases accuracy and increases Other Rate, suggesting a shift from biased answers toward uncertainty rather than correct grounding. Overall, RQ4 shows that PriVE-Tools does not identify a universally best visual evidence view; tool usefulness depends on both the counterfactual conflict and the model's ability to use localized evidence.

\section{Conclusion}
\label{sec:conclusion}

We introduced PriVE-Bench, a controlled counterfactual benchmark for distinguishing visually grounded answers from prior-consistent errors, and PriVE-Tools, an agentic-vision-inspired extension for evaluating whether tool-derived visual evidence helps VLMs reason against learned priors. Our experiments show that VLMs often follow language and category priors when counterfactual evidence contradicts canonical expectations. Paired images and tool-derived evidence can improve grounding in some settings, but their benefits are uneven across models, domains, and evidence types. Overall, our findings highlight a key challenge for agentic vision systems: exposing additional visual evidence does not guarantee that a VLM will use it to answer from what is actually visible.

\section{Limitations}
\label{sec:limitations}

PriVE-Tools is not a complete agentic vision framework. It evaluates controlled, pre-computed evidence views inspired by agentic systems, but does not measure adaptive tool selection, iterative inspection, planning, or self-correction. Some evidence is also partly oracle-like: boxes, crops, contours, and zoom panels are generated around known target regions, and medical evidence is derived from BraTS tumor segmentation labels. Our results therefore measure the usefulness of specific evidence views, not the end-to-end capability of autonomous visual agents.

Our tool coverage is not uniform across all domains and models. The main cross-domain comparison focuses on common evidence views, such as crops and zoom panels, while some domain-specific or partially completed conditions, such as bounding boxes, contours, and outlines, are reported separately. As a result, PriVE-Tools should be interpreted as a controlled diagnostic evaluation of selected evidence views rather than an exhaustive study of all possible visual tools.

Counterfactual construction may introduce artifacts. Non-medical examples can contain editing or generation artifacts, and AI-generated images may inherit the priors of the models used to create them. Web-collected images, in turn, may overlap with data seen during VLM pretraining. Although we apply filtering and quality control, we cannot fully rule out artifact-based behavior.

The medical-modality task is a controlled modality-consistency test rather than a clinical diagnosis task. Tumor-region signal is swapped between co-registered T1N and T2F slices to create a local contrast conflict, which is useful for testing visual grounding but does not simulate real MRI acquisition or pathology. The main medical evaluation also uses a single 2D axial slice and a stratified subset of the constructed pool, limiting coverage of full 3D radiological context.

Our model coverage is also limited. We evaluate representative closed- and open-source VLMs, but quota and cost constraints prevent a complete comparison across all current frontier systems. In particular, some models initially considered for evaluation could not be included in the formal matched comparison. Finally, yes/no and open-ended responses are partly scored by text-only LLM judges. Rubrics and deterministic multiple-choice scoring reduce ambiguity, but judge errors remain possible. The main setting also uses an explicitly prior-inducing prompt, making PriVE-Bench a stress test of visual grounding under salient priors rather than a complete characterization of all natural VLM interactions.

\bibliography{custom}

\begin{thebibliography}{30}
\providecommand{\natexlab}[1]{#1}

\bibitem[{Agrawal et~al.(2018)Agrawal, Batra, Parikh, and Kembhavi}]{agrawal2018don}
Aishwarya Agrawal, Dhruv Batra, Devi Parikh, and Aniruddha Kembhavi. 2018.
\newblock Don't just assume; look and answer: Overcoming priors for visual question answering.
\newblock In \emph{Proceedings of the IEEE conference on computer vision and pattern recognition}, pages 4971--4980.

\bibitem[{Alayrac et~al.(2022)Alayrac, Donahue, Luc, Miech, Barr, Hasson, Lenc, Mensch, Millican, Reynolds et~al.}]{alayrac2022flamingo}
Jean-Baptiste Alayrac, Jeff Donahue, Pauline Luc, Antoine Miech, Iain Barr, Yana Hasson, Karel Lenc, Arthur Mensch, Katherine Millican, Malcolm Reynolds, and 1 others. 2022.
\newblock Flamingo: a visual language model for few-shot learning.
\newblock \emph{Advances in neural information processing systems}, 35:23716--23736.

\bibitem[{Antol et~al.(2015)Antol, Agrawal, Lu, Mitchell, Batra, Zitnick, and Parikh}]{antol2015vqa}
Stanislaw Antol, Aishwarya Agrawal, Jiasen Lu, Margaret Mitchell, Dhruv Batra, C~Lawrence Zitnick, and Devi Parikh. 2015.
\newblock Vqa: Visual question answering.
\newblock In \emph{Proceedings of the IEEE international conference on computer vision}, pages 2425--2433.

\bibitem[{Bai et~al.(2025)Bai, Cai, Chen, Chen, Chen, Cheng, Deng, Ding, Gao, Ge et~al.}]{bai2025qwen3}
Shuai Bai, Yuxuan Cai, Ruizhe Chen, Keqin Chen, Xionghui Chen, Zesen Cheng, Lianghao Deng, Wei Ding, Chang Gao, Chunjiang Ge, and 1 others. 2025.
\newblock Qwen3-vl technical report.
\newblock \emph{arXiv preprint arXiv:2511.21631}.

\bibitem[{Baid et~al.(2021)Baid, Ghodasara, Mohan, Bilello, Calabrese, Colak, Farahani, Kalpathy-Cramer, Kitamura, Pati et~al.}]{baid2021rsna}
Ujjwal Baid, Satyam Ghodasara, Suyash Mohan, Michel Bilello, Evan Calabrese, Errol Colak, Keyvan Farahani, Jayashree Kalpathy-Cramer, Felipe~C Kitamura, Sarthak Pati, and 1 others. 2021.
\newblock The rsna-asnr-miccai brats 2021 benchmark on brain tumor segmentation and radiogenomic classification.
\newblock \emph{arXiv preprint arXiv:2107.02314}.

\bibitem[{Bitton-Guetta et~al.(2023)Bitton-Guetta, Bitton, Hessel, Schmidt, Elovici, Stanovsky, and Schwartz}]{bitton2023breaking}
Nitzan Bitton-Guetta, Yonatan Bitton, Jack Hessel, Ludwig Schmidt, Yuval Elovici, Gabriel Stanovsky, and Roy Schwartz. 2023.
\newblock Breaking common sense: Whoops! a vision-and-language benchmark of synthetic and compositional images.
\newblock In \emph{Proceedings of the IEEE/CVF International Conference on Computer Vision}, pages 2616--2627.

\bibitem[{Cadene et~al.(2019)Cadene, Dancette, Cord, Parikh et~al.}]{cadene2019rubi}
Remi Cadene, Corentin Dancette, Matthieu Cord, Devi Parikh, and 1 others. 2019.
\newblock Rubi: Reducing unimodal biases for visual question answering.
\newblock \emph{Advances in neural information processing systems}, 32.

\bibitem[{Chen et~al.(2024)Chen, Li, Dong, Zhang, Zang, Chen, Duan, Wang, Qiao, Lin et~al.}]{chen2024we}
Lin Chen, Jinsong Li, Xiaoyi Dong, Pan Zhang, Yuhang Zang, Zehui Chen, Haodong Duan, Jiaqi Wang, Yu~Qiao, Dahua Lin, and 1 others. 2024.
\newblock Are we on the right way for evaluating large vision-language models?
\newblock \emph{Advances in Neural Information Processing Systems}, 37:27056--27087.

\bibitem[{Deng et~al.(2025)Deng, Cao, Chen, and Hooi}]{deng2025words}
Ailin Deng, Tri Cao, Zhirui Chen, and Bryan Hooi. 2025.
\newblock Words or vision: Do vision-language models have blind faith in text?
\newblock In \emph{Proceedings of the Computer Vision and Pattern Recognition Conference}, pages 3867--3876.

\bibitem[{Guan et~al.(2024)Guan, Liu, Wu, Xian, Li, Liu, Wang, Chen, Huang, Yacoob et~al.}]{guan2024hallusionbench}
Tianrui Guan, Fuxiao Liu, Xiyang Wu, Ruiqi Xian, Zongxia Li, Xiaoyu Liu, Xijun Wang, Lichang Chen, Furong Huang, Yaser Yacoob, and 1 others. 2024.
\newblock Hallusionbench: an advanced diagnostic suite for entangled language hallucination and visual illusion in large vision-language models.
\newblock In \emph{Proceedings of the IEEE/CVF conference on computer vision and pattern recognition}, pages 14375--14385.

\bibitem[{Kamath et~al.(2025)Kamath, Ferret, Pathak, Vieillard, Merhej, Perrin, Matejovicova, Ram'e, Rivi{\`e}re, Rouillard, Mesnard, Cideron, Grill, Ramos, Yvinec, Casbon, Pot, Penchev, Liu, Visin, Kenealy, Beyer, Zhai, Tsitsulin, Busa-Fekete, Feng, Sachdeva, Coleman, Gao, Mustafa, Barr, Parisotto, Tian, Eyal, Cherry, Peter, Sinopalnikov, Bhupatiraju, Agarwal, Kazemi, Malkin, Kumar, Vilar, Brusilovsky, Luo, Steiner, Friesen, Sharma, Sharma, Gilady, Goedeckemeyer, Saade, Kolesnikov, Bendebury, Abdagic, Vadi, Gyorgy, Pinto, Das, Bapna, Miech, Yang, Paterson, Shenoy, Chakrabarti, Piot, Wu, Shahriari, Petrini, Chen, Lan, Choquette-Choo, Carey, Brick, Deutsch, Eisenbud, Cattle, Cheng, Paparas, Sreepathihalli, Reid, Tran, Zelle, Noland, Huizenga, Kharitonov, Liu, Amirkhanyan, Cameron, Hashemi, Klimczak-Pluci'nska, Singh, Mehta, Lehri, Hazimeh, Ballantyne, Szpektor, Nardini, Pouget-Abadie, Chan, Stanton, Wieting, Lai, Orbay, Fernandez, Newlan, Ji, Singh, Black, Yu, Hui, Vodrahalli, Greff,
  Qiu, Valentine, Coelho, Ritter, Hoffman, Watson, Chaturvedi, Moynihan, Ma, Babar, Noy, Byrd, Roy, Momchev, Chauhan, Bunyan, Botarda, Caron, Rubenstein, Culliton, Schmid, Sessa, mei Xu, Stańczyk, Tafti, Shivanna, Wu, Pan, Rokni, Willoughby, Vallu, Mullins, Jerome, Smoot, Girgin, Iqbal, Reddy, Sheth, P{\~o}der, Bhatnagar, Panyam, Eiger, Zhang, Liu, Yacovone, Liechty, Kalra, Evci, Misra, Roseberry, Feinberg, Kolesnikov, Han, Kwon, Chen, Chow, Zhu, Wei, Egyed, Cotruta, Giang, Kirk, Rao, Lo, Moreira, Martins, Sanseviero, Gonzalez, Gleicher, Warkentin, Mirrokni, Senter, Collins, Barral, Ghahramani, Hadsell, Matias, Sculley, Petrov, Fiedel, Shazeer, Vinyals, Dean, Hassabis, Kavukcuoglu, Farabet, Buchatskaya, Alayrac, Anil, Lepikhin, Borgeaud, Bachem, Joulin, Andreev, Hardin, Dadashi, and Hussenot}]{gemma2025gemma3}
Gemma Team~Aishwarya Kamath, Johan Ferret, Shreya Pathak, Nino Vieillard, Ramona Merhej, Sarah Perrin, Tatiana Matejovicova, Alexandre Ram'e, Morgane Rivi{\`e}re, Louis Rouillard, Thomas Mesnard, Geoffrey Cideron, Jean-Bastien Grill, Sabela Ramos, Edouard Yvinec, Michelle Casbon, Etienne Pot, Ivo Penchev, Gael Liu, and 191 others. 2025.
\newblock \href {https://api.semanticscholar.org/CorpusID:277313563} {Gemma 3 technical report}.
\newblock \emph{ArXiv}, abs/2503.19786.

\bibitem[{Kirillov et~al.(2023)Kirillov, Mintun, Ravi, Mao, Rolland, Gustafson, Xiao, Whitehead, Berg, Lo et~al.}]{kirillov2023segment}
Alexander Kirillov, Eric Mintun, Nikhila Ravi, Hanzi Mao, Chloe Rolland, Laura Gustafson, Tete Xiao, Spencer Whitehead, Alexander~C Berg, Wan-Yen Lo, and 1 others. 2023.
\newblock Segment anything.
\newblock In \emph{Proceedings of the IEEE/CVF international conference on computer vision}, pages 4015--4026.

\bibitem[{Lee et~al.(2025)Lee, Kim, Yoon, Kim, Lee, Koh, and Jung}]{lee2025vlind}
Kang-il Lee, Minbeom Kim, Seunghyun Yoon, Minsung Kim, Dongryeol Lee, Hyukhun Koh, and Kyomin Jung. 2025.
\newblock Vlind-bench: Measuring language priors in large vision-language models.
\newblock In \emph{Findings of the Association for Computational Linguistics: NAACL 2025}, pages 4129--4144.

\bibitem[{Li et~al.(2024)Li, Conte, Hu, Anwar, Kofler, Ezhov, van Leemput, Piraud, Diaz, Cole et~al.}]{li2024brain}
Hongwei~Bran Li, Gian~Marco Conte, Qingqiao Hu, Syed~Muhammad Anwar, Florian Kofler, Ivan Ezhov, Koen van Leemput, Marie Piraud, Maria Diaz, Byrone Cole, and 1 others. 2024.
\newblock The brain tumor segmentation (brats) challenge 2023: Brain mr image synthesis for tumor segmentation (brasyn).
\newblock \emph{ArXiv}, pages arXiv--2305.

\bibitem[{Li et~al.(2023)Li, Li, Savarese, and Hoi}]{li2023blip}
Junnan Li, Dongxu Li, Silvio Savarese, and Steven Hoi. 2023.
\newblock Blip-2: Bootstrapping language-image pre-training with frozen image encoders and large language models.
\newblock In \emph{International conference on machine learning}, pages 19730--19742. PMLR.

\bibitem[{Liu et~al.(2023{\natexlab{a}})Liu, Li, Wu, and Lee}]{liu2023visual}
Haotian Liu, Chunyuan Li, Qingyang Wu, and Yong~Jae Lee. 2023{\natexlab{a}}.
\newblock Visual instruction tuning.
\newblock \emph{Advances in neural information processing systems}, 36:34892--34916.

\bibitem[{Liu et~al.(2023{\natexlab{b}})Liu, Iter, Xu, Wang, Xu, and Zhu}]{liu2023g}
Yang Liu, Dan Iter, Yichong Xu, Shuohang Wang, Ruochen Xu, and Chenguang Zhu. 2023{\natexlab{b}}.
\newblock G-eval: Nlg evaluation using gpt-4 with better human alignment.
\newblock In \emph{Proceedings of the 2023 conference on empirical methods in natural language processing}, pages 2511--2522.

\bibitem[{Liu et~al.(2024)Liu, Dong, Rao, Zhou, and Lu}]{liu2024chain}
Zuyan Liu, Yuhao Dong, Yongming Rao, Jie Zhou, and Jiwen Lu. 2024.
\newblock Chain-of-spot: Interactive reasoning improves large vision-language models.
\newblock \emph{arXiv preprint arXiv:2403.12966}.

\bibitem[{Luo et~al.(2024)Luo, Cao, Lee, Johnson, and Lee}]{luo2024probing}
Tiange Luo, Ang Cao, Gunhee Lee, Justin Johnson, and Honglak Lee. 2024.
\newblock Probing visual language priors in vlms.
\newblock \emph{arXiv preprint arXiv:2501.00569}.

\bibitem[{Menze et~al.(2014)Menze, Jakab, Bauer, Kalpathy-Cramer, Farahani, Kirby, Burren, Porz, Slotboom, Wiest et~al.}]{menze2014multimodal}
Bjoern~H Menze, Andras Jakab, Stefan Bauer, Jayashree Kalpathy-Cramer, Keyvan Farahani, Justin Kirby, Yuliya Burren, Nicole Porz, Johannes Slotboom, Roland Wiest, and 1 others. 2014.
\newblock The multimodal brain tumor image segmentation benchmark (brats).
\newblock \emph{IEEE transactions on medical imaging}, 34(10):1993--2024.

\bibitem[{Radford et~al.(2021)Radford, Kim, Hallacy, Ramesh, Goh, Agarwal, Sastry, Askell, Mishkin, Clark et~al.}]{radford2021learning}
Alec Radford, Jong~Wook Kim, Chris Hallacy, Aditya Ramesh, Gabriel Goh, Sandhini Agarwal, Girish Sastry, Amanda Askell, Pamela Mishkin, Jack Clark, and 1 others. 2021.
\newblock Learning transferable visual models from natural language supervision.
\newblock In \emph{International conference on machine learning}, pages 8748--8763. PmLR.

\bibitem[{Vo et~al.(2025)Vo, Nguyen, Taesiri, Dang, Nguyen, and Kim}]{vo2025vision}
An~Vo, Khai-Nguyen Nguyen, Mohammad~Reza Taesiri, Vy~Tuong Dang, Anh~Totti Nguyen, and Daeyoung Kim. 2025.
\newblock Vision language models are biased.
\newblock \emph{arXiv preprint arXiv:2505.23941}.

\bibitem[{Wang et~al.(2025)Wang, Gao, Gu, Pu, Cui, Wei, Liu, Jing, Ye, Shao et~al.}]{wang2025internvl3}
Weiyun Wang, Zhangwei Gao, Lixin Gu, Hengjun Pu, Long Cui, Xingguang Wei, Zhaoyang Liu, Linglin Jing, Shenglong Ye, Jie Shao, and 1 others. 2025.
\newblock Internvl3.5: Advancing open-source multimodal models in versatility, reasoning, and efficiency.
\newblock \emph{arXiv preprint arXiv:2508.18265}.

\bibitem[{Wu et~al.(2024)Wu, Zhang, Xia, Li, Xia, Chang, Yu, Kim, Rossi, Zhang et~al.}]{wu2024visual}
Junda Wu, Zhehao Zhang, Yu~Xia, Xintong Li, Zhaoyang Xia, Aaron Chang, Tong Yu, Sungchul Kim, Ryan~A Rossi, Ruiyi Zhang, and 1 others. 2024.
\newblock Visual prompting in multimodal large language models: A survey.
\newblock \emph{arXiv preprint arXiv:2409.15310}.

\bibitem[{Wu and Xie(2024)}]{wu2024v}
Penghao Wu and Saining Xie. 2024.
\newblock V?: Guided visual search as a core mechanism in multimodal llms.
\newblock In \emph{Proceedings of the IEEE/CVF Conference on Computer Vision and Pattern Recognition}, pages 13084--13094.

\bibitem[{Yang et~al.(2023{\natexlab{a}})Yang, Zhang, Li, Zou, Li, and Gao}]{yang2023set}
Jianwei Yang, Hao Zhang, Feng Li, Xueyan Zou, Chunyuan Li, and Jianfeng Gao. 2023{\natexlab{a}}.
\newblock Set-of-mark prompting unleashes extraordinary visual grounding in gpt-4v.
\newblock \emph{arXiv preprint arXiv:2310.11441}.

\bibitem[{Yang et~al.(2023{\natexlab{b}})Yang, Li, Wang, Lin, Azarnasab, Ahmed, Liu, Liu, Zeng, and Wang}]{yang2023mm}
Zhengyuan Yang, Linjie Li, Jianfeng Wang, Kevin Lin, Ehsan Azarnasab, Faisal Ahmed, Zicheng Liu, Ce~Liu, Michael Zeng, and Lijuan Wang. 2023{\natexlab{b}}.
\newblock Mm-react: Prompting chatgpt for multimodal reasoning and action.
\newblock \emph{arXiv preprint arXiv:2303.11381}.

\bibitem[{Yao et~al.(2022)Yao, Zhao, Yu, Du, Shafran, Narasimhan, and Cao}]{yao2022react}
Shunyu Yao, Jeffrey Zhao, Dian Yu, Nan Du, Izhak Shafran, Karthik Narasimhan, and Yuan Cao. 2022.
\newblock React: Synergizing reasoning and acting in language models.
\newblock \emph{arXiv preprint arXiv:2210.03629}.

\bibitem[{Zhang et~al.(2023)Zhang, Khayatkhoei, Chhikara, and Ilievski}]{zhang2023towards}
Jiarui Zhang, Mahyar Khayatkhoei, Prateek Chhikara, and Filip Ilievski. 2023.
\newblock Towards perceiving small visual details in zero-shot visual question answering with multimodal llms.
\newblock \emph{arXiv preprint arXiv:2310.16033}.

\bibitem[{Zheng et~al.(2023)Zheng, Chiang, Sheng, Zhuang, Wu, Zhuang, Lin, Li, Li, Xing et~al.}]{zheng2023judging}
Lianmin Zheng, Wei-Lin Chiang, Ying Sheng, Siyuan Zhuang, Zhanghao Wu, Yonghao Zhuang, Zi~Lin, Zhuohan Li, Dacheng Li, Eric Xing, and 1 others. 2023.
\newblock Judging llm-as-a-judge with mt-bench and chatbot arena.
\newblock \emph{Advances in neural information processing systems}, 36:46595--46623.

\end{thebibliography}

\clearpage
\appendix
\twocolumn[
{\LARGE\bfseries Appendix}\par
\vspace{1em}
]

\section{Dataset Statistics and Source Composition}
\label{app:dataset_statistics}

Appendix~\ref{app:dataset_statistics} summarizes the dataset composition used in our formal evaluation. For non-medical domains, \emph{Originals} denotes unique original images, while \emph{CF pairs} denotes active counterfactual evaluation records. These counts are not always identical: a single original image may support multiple counterfactual interventions, such as modifying a logo layout, changing an object color, removing a visual attribute, or altering a countable part.

We also report the source composition of original images. The non-medical domains combine web-collected or real images with AI-generated images. This mixed construction is intentional. Web-collected images preserve natural visual diversity and realistic object appearances, but may overlap with data seen during VLM pretraining. AI-generated images provide a more controllable source for constructing targeted counterfactual attributes, but may themselves reflect the visual priors of the image-generation models used to create them. We therefore record source metadata and report the AI-generated proportion for each non-medical domain, rather than treating either source type as fully unbiased or guaranteed unseen.

\begin{figure*}[t!]
    \centering
    \includegraphics[width=\textwidth]{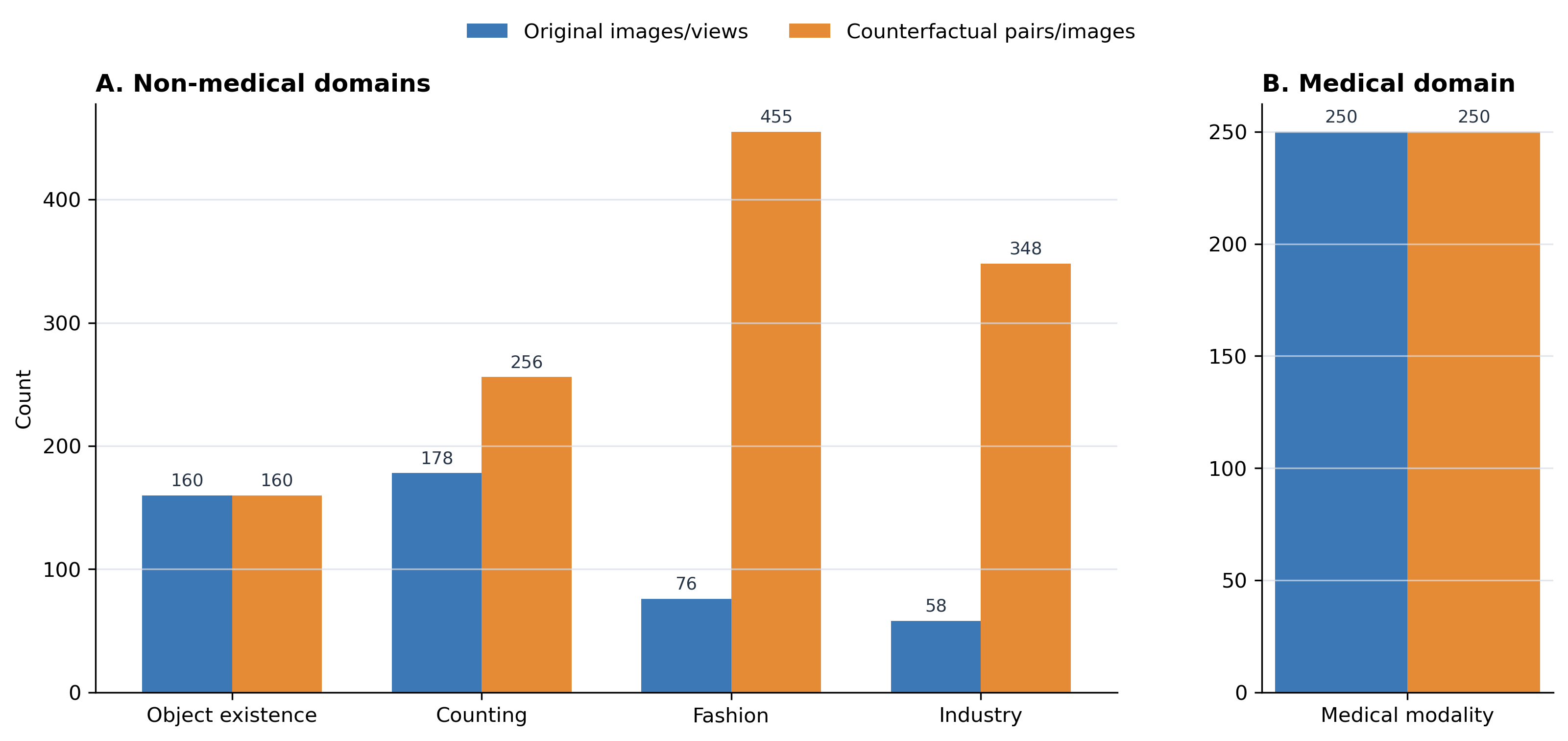}
    \caption{
    \textbf{Original and counterfactual counts by domain in the formal evaluation set.} Counts are computed from active benchmark metadata. Counting reuses some originals across multiple edits, while medical modality reports the stratified evaluation subset; the full constructed medical pool contains 2,502 counterfactual modality-swap records.
    }
    \label{fig:app_counts}
\end{figure*}

\begin{figure*}[t!]
    \centering
    \includegraphics[width=\textwidth]{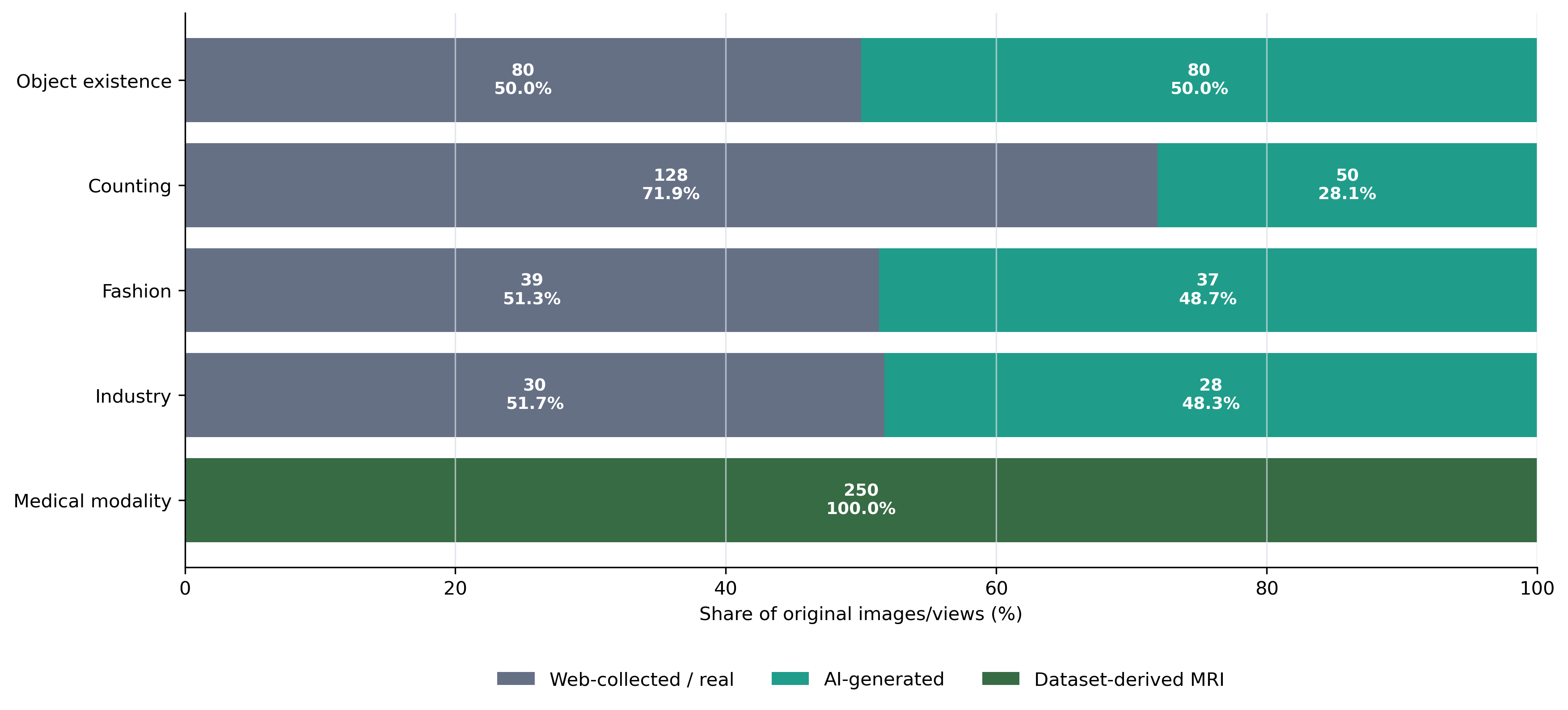}
    \caption{
    \textbf{Source composition of original images.} Non-medical domains combine web-collected or real images with AI-generated images to balance natural visual diversity and controllable counterfactual construction. Medical examples are derived from BraTS2023 MRI data and are reported separately as dataset-derived medical images.
    }
    \label{fig:app_sources}
\end{figure*}

The medical-modality domain is handled separately. It is derived from the ASNR-MICCAI-BraTS2023-GLI training data, following the BraTS multimodal brain tumor segmentation benchmark setting~\citep{menze2014multimodal,baid2021rsna}. Although we construct a full clean pool of modality-swapped counterfactual examples, the main experiments use a stratified 10\% subset to reduce evaluation cost while preserving coverage across cases and swap directions. Therefore, Table~\ref{tab:dataset_statistics} reports both the formal evaluation subset and the full constructed counterfactual pool. Figures~\ref{fig:app_counts} and~\ref{fig:app_sources} visualize the same metadata statistics.

\begin{table*}[t!]
\centering
\small
\setlength{\tabcolsep}{4pt}
\begin{tabular*}{\textwidth}{@{\extracolsep{\fill}}lrrrp{0.24\textwidth}p{0.24\textwidth}@{}}
\toprule
\textbf{Domain} &
\textbf{Orig.} &
\textbf{Eval. CF} &
\textbf{Full CF} &
\textbf{Original source} &
\textbf{Notes} \\
\midrule
Object existence &
160 & 160 & 160 &
80 web / 80 AI (50.0\%) &
Expected object parts removed \\

Counting &
178 & 256 & 256 &
128 web / 50 AI (28.1\%) &
Some originals reused across edits \\

Fashion &
76 & 455 & 455 &
39 web / 37 AI (48.7\%) &
Filtered logo and monogram edits \\

Industry &
58 & 348 & 348 &
30 web / 28 AI (48.3\%) &
Filtered common-object edits \\

Medical modality &
250 & 250 & 2,502 &
250 BraTS MRI &
Stratified 10\% subset of modality swaps \\
\midrule
\textbf{Total} &
\textbf{722} & \textbf{1,469} & \textbf{3,721} &
-- &
-- \\
\bottomrule
\end{tabular*}
\caption{
\textbf{Dataset composition used in PriVE-Bench.} Counts are computed from active benchmark metadata rather than raw directory scans. \emph{Orig.} denotes unique original images for non-medical domains and standardized MRI source views for the medical evaluation subset. \emph{Eval. CF} denotes counterfactual records used in formal evaluation, while \emph{Full CF} denotes all constructed counterfactual records before medical subsampling. AI percentages are computed only for non-medical originals.
}
\label{tab:dataset_statistics}
\end{table*}

\section{Medical Modality Sampling and Counterfactual Construction}
\label{app:medical_sampling}

\paragraph{Data source.}
The medical-modality subset of PriVE-Bench is constructed from the ASNR-MICCAI-BraTS2023-GLI Challenge training data, following the BraTS multimodal brain tumor segmentation benchmark setting~\citep{menze2014multimodal,baid2021rsna,li2024brain}. Each case contains co-registered multi-modal brain MRI volumes and a dense tumor segmentation label. We use the native T1-weighted MRI (T1N), T2-FLAIR MRI (T2F), and the provided segmentation mask. The segmentation mask is used only to define the tumor region for controlled counterfactual construction and, in PriVE-Tools, to generate localization-based visual evidence. It is not shown to the model as a semantic tumor label in the raw PriVE-Bench setting.

Unlike the non-medical domains, the medical subset is not based on web-collected or text-to-image generated pictures. It is derived from real multi-modal MRI data in which different imaging sequences are spatially aligned for the same patient case. This property makes BraTS suitable for controlled modality-consistency counterfactuals: local tumor-region signal can be exchanged between modalities while preserving patient identity, slice location, tumor shape, and surrounding anatomy.

\paragraph{Slice selection and standardized visualization.}
For each patient case, we select one axial slice for evaluation. The selected slice is the slice with the largest whole-tumor area, where the whole-tumor mask is defined as all non-zero BraTS segmentation labels. This avoids using slices in which the tumor is absent or too small to support reliable visual judgment.

For the selected slice, we extract the corresponding T1N and T2F images. Each slice is normalized using the non-zero brain region, with intensities clipped and scaled by robust percentile statistics, and saved as a grayscale PNG. The main benchmark uses the \textsc{clean} visualization version: neither the original nor the counterfactual image contains a colored tumor overlay. This ensures that the task tests whether the model can infer modality consistency from visible MRI signal, rather than exploiting an explicit mask or annotation.

\paragraph{Counterfactual modality swap.}
The core intervention is a local tumor-region modality swap. Let $I^{\mathrm{T1N}}$ and $I^{\mathrm{T2F}}$ denote the aligned T1N and T2F slices from the same patient case, and let $M$ be the binary whole-tumor mask for the selected slice. We construct two directional counterfactuals:
\begin{align}
C^{\mathrm{T1N}\leftarrow\mathrm{T2F}} &= (1-M) \odot I^{\mathrm{T1N}} + M \odot I^{\mathrm{T2F}}, \\
C^{\mathrm{T2F}\leftarrow\mathrm{T1N}} &= (1-M) \odot I^{\mathrm{T2F}} + M \odot I^{\mathrm{T1N}}.
\end{align}
In the first direction, the image retains the global T1N background, but pixels inside the tumor mask are replaced with the corresponding T2F tumor-region pixels. In the second direction, the image retains the global T2F background, but the tumor region is replaced with the corresponding T1N tumor-region pixels. Because the modalities are co-registered, the replacement is spatially aligned and does not require an external generative model.

\paragraph{Imaging-sequence rationale.}
This intervention creates a controlled conflict between global modality appearance and local tumor-region contrast. In a physically consistent MRI slice, the tumor region and surrounding tissue are acquired under the same imaging sequence, even though their intensities may differ due to tissue properties. Our counterfactual images deliberately violate this consistency: the surrounding brain tissue indicates one MRI sequence, while the tumor region carries the local contrast pattern of another sequence.

The task is therefore not clinical diagnosis, tumor typing, or a realistic simulation of MRI acquisition. It is a controlled visual counterfactual for testing modality-consistency judgment. A visually grounded model should detect that the tumor-region contrast conflicts with the surrounding sequence. A prior-following model may instead answer according to the dominant global modality appearance and assume that the tumor region is consistent with the rest of the slice.

\paragraph{Sampling protocol.}
The full constructed clean pool contains 1,251 BraTS2023 cases and two modality-swap directions per case, yielding 2,502 counterfactual records. For the main evaluation, we use a stratified 10\% subset to reduce evaluation cost while preserving balance across swap directions.

Sampling is performed before question expansion. The sampling strata are defined by dataset, visualization version, and swap direction. In the main setting, all samples come from BraTS2023 with the \textsc{clean} visualization, so the two effective strata correspond to the two swap directions: T1N-background/T2F-tumor and T2F-background/T1N-tumor. We sample 125 records from each direction, yielding 250 medical counterfactual records for formal evaluation. The random seed is fixed in the evaluation code, and the selected pair keys are recorded in the run configuration through a digest, enabling runs to be audited and reproduced.

\paragraph{Label-derived visual evidence for PriVE-Tools.}
For PriVE-Tools, the medical subset uses label-derived visual evidence rather than natural-image segmentation outputs. This avoids applying a generic segmentation model to MRI images, where domain mismatch could introduce unstable or clinically meaningless masks. The BraTS tumor segmentation label provides a consistent localization source for constructing controlled evidence views.

From the same whole-tumor mask, we generate four medical evidence conditions: bounding-box overlay, contour overlay, tumor crop, and zoom panel. The bounding box and contour provide thin localization cues without filling or recoloring the tumor region. The crop and zoom panel magnify the same tumor-region pixels while preserving the raw image context as part of the model input. We exclude filled mask overlays and difference maps from the main experiments because they could either obscure the tumor intensity pattern or directly reveal the intervention, turning the task into mask interpretation rather than visual evidence grounding.

\section{PriVE-Tools Evidence Conditions}
\label{app:tool_conditions}

PriVE-Tools evaluates controlled tool-derived evidence conditions inspired by common operations in agentic or tool-augmented vision systems. These conditions modify how the relevant visual region is presented to the model while keeping the underlying question, target attribute, correct answer, biased answer, and scoring rubric fixed. They do not include textual labels, edit descriptions, arrows indicating the manipulation, or explicit answer cues.

\begin{table}[t!]
\centering
\small
\begin{tabular*}{\linewidth}{@{\extracolsep{\fill}}p{0.24\linewidth}p{0.68\linewidth}@{}}
\toprule
\textbf{Condition} & \textbf{Purpose} \\
\midrule
BBox & Localizes the target object or region without changing image content. \\
Crop & Magnifies the relevant region and reduces irrelevant context. \\
Zoom panel & Preserves the full image while adding an enlarged local view. \\
Contour & Highlights a boundary without filling or recoloring the region. \\
Tool bundle & Provides multiple complementary evidence views together. \\
\bottomrule
\end{tabular*}
\caption{
\textbf{Tool-derived evidence conditions in PriVE-Tools.} Available conditions vary by domain according to the target attribute and evidence source.
}
\end{table}

For natural-image domains, evidence is generated from object-level localization or segmentation outputs and manually filtered where necessary. Counting examples may use contour-style evidence to highlight countable instances. Medical-modality examples use label-derived evidence from the BraTS tumor segmentation mask, avoiding generic natural-image segmentation models on MRI data. We exclude filled masks and difference maps from the main evaluation because they may obscure the relevant visual signal or directly reveal the counterfactual manipulation.

\section{Evaluation Configuration}
\label{app:evaluation_config}

Table~\ref{tab:evaluation_config} summarizes the default configuration used for formal PriVE-Bench and PriVE-Tools evaluations. These settings are chosen to reduce decoding randomness, standardize judge behavior, and make raw-image and tool-evidence comparisons as controlled as possible. Unless otherwise stated, provider-specific fallbacks are recorded in the run metadata.

\begin{table*}[t!]
\centering
\small
\setlength{\tabcolsep}{5pt}
\begin{tabular*}{\textwidth}{@{\extracolsep{\fill}}p{0.22\textwidth}p{0.20\textwidth}p{0.50\textwidth}@{}}
\toprule
\textbf{Setting} & \textbf{Default} & \textbf{Rationale} \\
\midrule
Temperature & 0 & Uses deterministic decoding where supported, reducing run-to-run variation. \\

Prior-conflict prime & On & Makes the relevant canonical prior explicit and consistent across models in the main evaluation setting. \\

Backbone max output tokens & 2048 & Provides enough budget for short answers, formatting, and provider overhead without encouraging unrestricted long-form responses. \\

Text-only judge & GPT-4o-mini for closed-source VLM runs; Qwen3-8B for open-source VLM runs & Uses a fixed text-only judge within each model group for rubric-based scoring of yes/no and open-ended responses. The judge receives the question, model response, and correct/biased rubrics, but not the image. \\

Judge max output tokens & 512 & The judge only needs to return a label and short rationale, so a smaller budget is sufficient. \\

Judge structured output & Auto & Requests JSON-style judge labels where supported, with fallback parsing for unsupported providers. \\

Closed-form structured output & Auto & Requests constrained outputs for yes/no and multiple-choice questions where supported. \\

Reasoning or extended thinking & Off; minimal only if required & Avoids enabling model-specific deliberation as an uncontrolled advantage. \\

Missing evidence policy & Skip & Tool-condition examples without required evidence are skipped before inference. \\

Error handling & Excluded from denominator; reported separately & API, server, quota, authentication, and judge failures are not treated as model answers. \\
\bottomrule
\end{tabular*}
\caption{
\textbf{Default evaluation configuration for formal PriVE-Bench and PriVE-Tools runs.} Provider-specific deviations and fallbacks are recorded in run metadata. The text-only judge is used only to classify completed model responses according to the predefined correct/biased/other rubrics.
}
\label{tab:evaluation_config}
\end{table*}

\paragraph{Decoding and output budgets.}
Formal runs use temperature 0 wherever supported. When a provider does not expose exact deterministic decoding, we use the closest supported configuration and record the provider-specific settings. The backbone output budget is set to 2048 tokens. Although most benchmark answers are short, this budget reduces empty-output and output-limit failures for models that prepend formatting, safety text, or brief explanations. The judge budget is set to 512 tokens because the judge is only required to output a label and a short rationale.

\paragraph{Prior-conflict prompting.}
The prior-conflict prime is enabled in the main formal evaluation. PriVE-Bench is designed to test whether models can ground their answers in visible evidence when a relevant language or category prior is explicitly salient. Enabling the same prior-inducing context across models makes the conflict controlled rather than relying on implicit prompt effects. Runs with different prompt policies are treated as separate configurations.

\paragraph{Structured outputs and reasoning controls.}
Structured output is requested automatically where supported. For judged questions, structured output makes the judge label easier to parse and audit. For closed-form questions, it encourages models to return a constrained yes/no answer or option letter. If a provider does not support structured output, the runner falls back to text parsing and records the fallback reason. Reasoning or extended-thinking controls are disabled by default, because the benchmark is intended to compare visual grounding under the same input evidence rather than compare different levels of model-specific deliberation. If a provider requires a minimal reasoning setting, we use the minimum supported mode and record it.

\paragraph{Missing evidence and matched comparisons.}
For PriVE-Tools, missing evidence is handled with a skip policy. If a required crop, bounding box, contour, outline, or zoom panel is unavailable, the example is not sent to the model. This avoids mixing raw-image behavior with incomplete tool-input behavior. When raw and tool runs have different eligible subsets, tool deltas should either report the denominator difference or be computed on the matched evidence-valid subset.

\paragraph{Error handling.}
Infrastructure and provider failures are excluded from the metric denominator and reported separately. These include API quota errors, authentication failures, server startup failures, network errors, empty responses after retry, and judge failures. Such events are not merged into the \emph{other} category, because \emph{other} is reserved for completed model responses that are ambiguous, off-target, contradictory, or unparseable with respect to the task rubric.

\section{Full Raw PriVE-Bench Results}
\label{app:raw_results}

This appendix reports the raw-image results supporting RQ1. We focus on the two input modes most directly relevant to prior-following behavior: \textsc{original-only} and \textsc{counterfactual-only}. The original-only setting checks whether the unedited images and questions are answerable, while the counterfactual-only setting tests whether models answer from visible evidence when the image violates a canonical expectation.

All metrics are computed over valid, successfully scored model responses. Query errors and judge errors are excluded from the metric denominator and reported separately in experiment logs. When runs are resumed after transient failures, duplicate task signatures are resolved by retaining the successful record when available. Unless otherwise stated, values are percentages.

\paragraph{Counterfactual-only results.}
Table~\ref{tab:raw_cf_by_domain} reports raw counterfactual-only performance by model and domain. Each cell reports Accuracy / PFER / Other Rate. Accuracy measures visually correct responses, PFER measures prior-consistent errors, and Other Rate captures ambiguous, noncommittal, or otherwise uncategorizable responses.

\begin{table*}[t]
\centering
\scriptsize
\setlength{\tabcolsep}{3.2pt}
\begin{tabular*}{\textwidth}{@{\extracolsep{\fill}}lccccc@{}}
\toprule
\textbf{Model} & 
\textbf{Existence} & 
\textbf{Counting} & 
\textbf{Fashion} & 
\textbf{Industry} & 
\textbf{Medical} \\
\midrule
GPT-5 
& 57.7/36.5/5.8 
& 15.1/84.5/0.4 
& 44.3/54.5/1.2 
& 85.1/14.5/0.4 
& 31.5/66.3/2.3 \\

Claude Sonnet 4 
& 62.3/36.2/1.5 
& 19.5/73.7/6.8 
& 42.6/56.0/1.3 
& 67.3/32.3/0.4 
& 41.5/56.8/1.7 \\

Qwen3-VL-8B-Instruct 
& 31.9/62.1/6.0 
& 14.8/83.2/2.0 
& 32.5/65.7/1.8 
& 51.7/48.1/0.3 
& 27.2/72.1/0.7 \\

Qwen3-VL-32B-Instruct 
& 74.4/23.1/2.5 
& 13.0/85.8/1.2 
& 27.3/72.6/0.1 
& 53.6/46.1/0.3 
& 33.1/66.4/0.5 \\

InternVL3.5-8B 
& 31.0/65.6/3.3 
& 5.1/94.1/0.8 
& 25.2/73.1/1.7 
& 56.4/43.2/0.4 
& 19.9/67.2/12.9 \\

InternVL3.5-38B 
& 53.1/45.8/1.0 
& 4.0/85.0/10.9 
& 27.0/71.9/1.1 
& 65.9/33.7/0.4 
& 37.7/62.3/0.0 \\

Gemma-3-12B-it 
& 46.7/44.6/8.8 
& 36.8/52.7/10.4 
& 18.7/74.4/6.8 
& 29.7/68.4/1.9 
& 66.8/29.3/3.9 \\

Gemma-3-27B-it 
& 65.6/32.5/1.9 
& 27.6/63.9/8.5 
& 18.5/78.7/2.9 
& 21.6/77.4/1.0 
& 19.2/53.1/27.7 \\
\bottomrule
\end{tabular*}
\caption{
\textbf{Raw counterfactual-only results by model and domain.} Each cell reports Acc/PFER/Other, in percent. High PFER indicates prior-consistent errors under counterfactual visual evidence.
}
\label{tab:raw_cf_by_domain}
\end{table*}

Table~\ref{tab:raw_cf_by_domain} shows that counterfactual failures are often not random uncertainty. Across many models and domains, PFER is substantially larger than Other Rate, indicating that errors frequently preserve the learned prior rather than avoid commitment. Counting and fashion are especially strong stress tests, suggesting that canonical expectations about object counts and visual-design patterns can override edited visual evidence.

\paragraph{Original-only sanity check.}
Table~\ref{tab:orig_sanity_by_domain} reports original-only accuracy by model and domain. These results verify that the unedited images and question templates are generally answerable before counterfactual interventions are introduced.

\begin{table*}[t]
\centering
\scriptsize
\setlength{\tabcolsep}{5pt}
\begin{tabular*}{\textwidth}{@{\extracolsep{\fill}}lccccc@{}}
\toprule
\textbf{Model} & 
\textbf{Existence} & 
\textbf{Counting} & 
\textbf{Fashion} & 
\textbf{Industry} & 
\textbf{Medical} \\
\midrule
GPT-5 & 97.1 & 93.2 & 67.8 & 72.6 & 60.8 \\
Claude Sonnet 4 & 88.8 & 84.1 & 55.5 & 77.7 & 47.6 \\
Qwen3-VL-8B-Instruct & 99.6 & 88.3 & 67.9 & 93.4 & 57.2 \\
Qwen3-VL-32B-Instruct & 94.6 & 94.4 & 77.7 & 88.1 & 63.5 \\
InternVL3.5-8B & 68.5 & 95.8 & 69.6 & 92.7 & 72.4 \\
InternVL3.5-38B & 81.2 & 85.5 & 62.6 & 88.0 & 48.4 \\
Gemma-3-12B-it & 69.6 & 54.0 & 63.1 & 86.9 & 31.1 \\
Gemma-3-27B-it & 73.8 & 69.8 & 64.9 & 86.4 & 46.1 \\
\bottomrule
\end{tabular*}
\caption{
\textbf{Original-only sanity-check accuracy by model and domain, in percent.} High original-only accuracy indicates that the unedited images and questions are generally answerable.
}
\label{tab:orig_sanity_by_domain}
\end{table*}

The original-only results are generally higher than the corresponding counterfactual-only results, supporting the interpretation that many counterfactual failures are not solely due to unanswerable questions or poor baseline recognition. The medical domain is more difficult even in the original-only setting, so medical results should be interpreted as a stronger stress test of local modality-consistency reasoning rather than as a standard recognition task.

\paragraph{Domain-level summary.}
Table~\ref{tab:rq1_domain_macro} reports domain-level macro-averages over models. This table supports the domain-level analysis in Section~\ref{sec:experiments}. Counting shows the largest original-to-counterfactual drop and the highest PFER, while fashion also produces strong prior-following behavior. Medical modality consistency has lower original-only accuracy, but still shows substantial PFER under counterfactual inputs.

\begin{table*}[t]
\centering
\small
\setlength{\tabcolsep}{4pt}
\begin{tabular*}{\textwidth}{@{\extracolsep{\fill}}lrrrrr@{}}
\toprule
\textbf{Domain} &
\textbf{Orig. Acc} &
\textbf{CF Acc} &
\textbf{CF PFER} &
\textbf{CF Other} &
\textbf{Drop} \\
\midrule
Existence & 84.2 & 52.8 & 43.3 & 3.9 & 31.3 \\
Counting & 83.1 & 17.0 & 77.9 & 5.1 & 66.2 \\
Fashion & 66.1 & 29.5 & 68.4 & 2.1 & 36.6 \\
Industry & 85.7 & 53.9 & 45.5 & 0.6 & 31.8 \\
Medical & 53.4 & 34.6 & 59.2 & 6.2 & 18.8 \\
\bottomrule
\end{tabular*}
\caption{
\textbf{Domain-level macro-averages over models for RQ1.} \textbf{Drop} is the difference between original-only accuracy and counterfactual-only accuracy. Counting and fashion are the strongest prior-following stress tests, while medical modality consistency remains challenging even in the original-only setting.
}
\label{tab:rq1_domain_macro}
\end{table*}

\begin{figure*}[t]
    \centering
    \includegraphics[width=\textwidth]{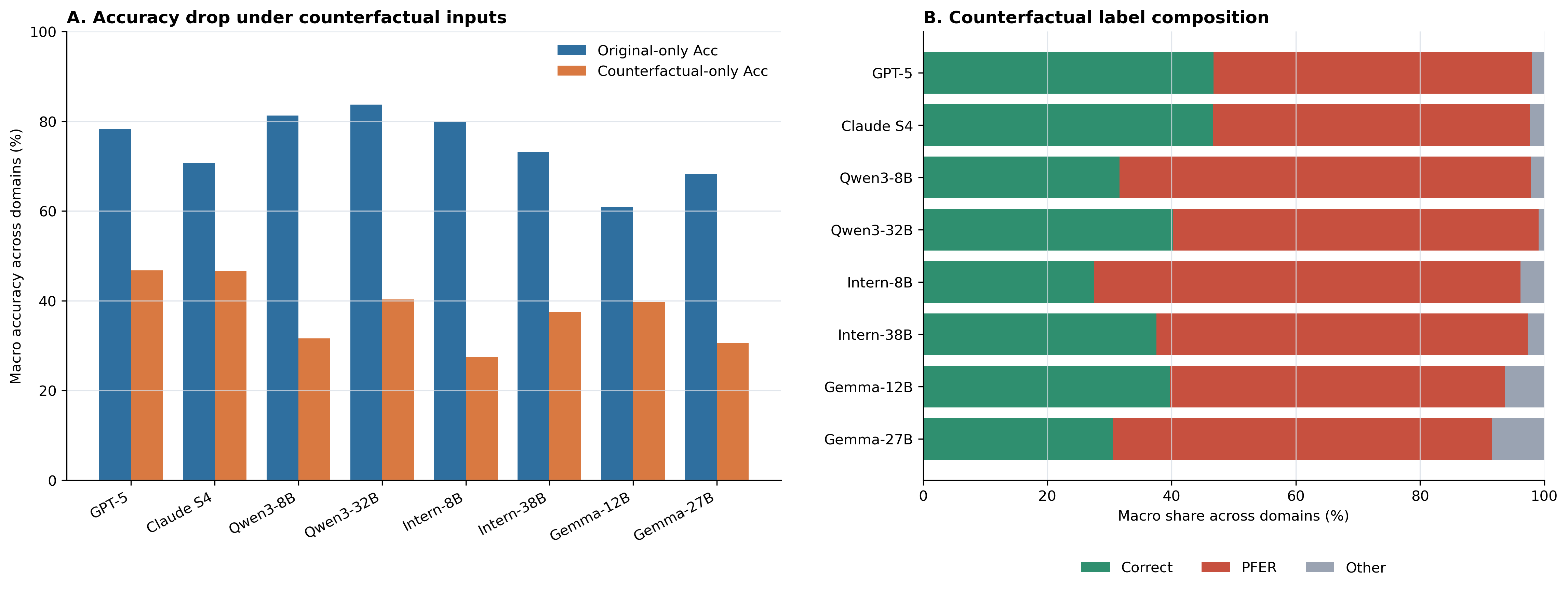}
    \caption{
    \textbf{Original-only and counterfactual-only raw-image performance.} 
    Panel A compares answerability under original images with performance under counterfactual images. 
    Panel B decomposes counterfactual-only responses into correct, prior-following, and other labels. 
    The large PFER component shows that counterfactual failures are often prior-consistent rather than merely ambiguous.
    }
    \label{fig:orig_vs_cf_accuracy}
\end{figure*}

\paragraph{Original versus counterfactual performance.}
Figure~\ref{fig:orig_vs_cf_accuracy} summarizes the model-level pattern. Panel A compares macro-averaged original-only and counterfactual-only accuracy. Panel B decomposes counterfactual-only responses into correct, prior-following, and other labels.

Together, these results provide the raw evidence for RQ1. Original-only performance establishes baseline answerability, while counterfactual-only performance reveals a systematic shift toward prior-consistent errors. This supports the conclusion that current VLMs often fall back on learned category or modality priors when visible evidence contradicts what is usually true.

\section{Paired-Image Analysis}
\label{app:paired_results}

This appendix reports the full paired-image results used to support RQ2. In the paired-image setting, each model receives both the original and counterfactual image with neutral labels and is asked to compare the target visual attribute. This setting tests whether direct visual comparison helps models detect the counterfactual intervention, rather than answering from the prior associated with the object, category, or modality.

All metrics are computed over valid, successfully scored responses. Query and judge errors are excluded from metric denominators and tracked separately in experiment logs. Unless otherwise stated, values are percentages for absolute results and percentage points for deltas.

\paragraph{Absolute paired-image results.}
Table~\ref{tab:paired_by_domain} reports paired-image performance by model and domain. Each cell reports Accuracy / PFER / Other Rate. Accuracy measures visually correct answers, PFER measures prior-consistent but visually incorrect answers, and Other Rate captures ambiguous, noncommittal, or otherwise uncategorizable responses.

\begin{table*}[t]
\centering
\scriptsize
\setlength{\tabcolsep}{3.2pt}
\begin{tabular*}{\textwidth}{@{\extracolsep{\fill}}lccccc@{}}
\toprule
\textbf{Model} & 
\textbf{Existence} & 
\textbf{Counting} & 
\textbf{Fashion} & 
\textbf{Industry} & 
\textbf{Medical} \\
\midrule
GPT-5 
& 65.0/25.6/9.4 
& 26.7/72.7/0.7 
& 54.8/44.7/0.4 
& 86.3/13.2/0.6 
& 39.9/58.8/1.3 \\

Claude Sonnet 4 
& 54.2/35.0/10.8 
& 28.5/68.1/3.4 
& 42.3/57.6/0.2 
& 75.7/24.2/0.1 
& 36.1/62.0/1.9 \\

Qwen3-VL-8B-Instruct 
& 20.6/67.9/11.5 
& 14.3/59.6/26.0 
& 36.4/63.6/0.0 
& 72.9/27.1/0.0 
& 64.4/19.7/15.9 \\

Qwen3-VL-32B-Instruct 
& 57.5/31.5/11.0 
& 35.9/45.1/19.0 
& 35.1/64.9/0.0 
& 67.3/32.6/0.1 
& 52.9/46.8/0.3 \\

InternVL3.5-8B 
& 1.2/88.3/10.4 
& 32.7/64.2/3.1 
& 32.3/67.4/0.3 
& 66.1/33.9/0.0 
& 27.1/72.9/0.0 \\

InternVL3.5-38B 
& 26.5/69.0/4.6 
& 32.6/57.7/9.8 
& 39.5/60.5/0.0 
& 71.0/28.9/0.1 
& 38.4/61.6/0.0 \\

Gemma-3-12B-it 
& 27.5/61.9/10.6 
& 29.9/43.5/26.6 
& 33.6/66.4/0.0 
& 31.3/68.7/0.0 
& 36.4/63.5/0.1 \\

Gemma-3-27B-it 
& 30.8/57.1/12.1 
& 32.0/49.5/18.5 
& 41.2/58.7/0.1 
& 61.4/38.6/0.0 
& 44.0/54.7/1.3 \\
\bottomrule
\end{tabular*}
\caption{
\textbf{Paired-image results by model and domain.} Each cell reports Acc/PFER/Other, in percent, when models compare original and counterfactual images with neutral labels.
}
\label{tab:paired_by_domain}
\end{table*}

Table~\ref{tab:paired_by_domain} shows that paired-image inputs do not uniformly solve prior-following behavior. In several domains, models still produce high PFER even when the original image is available for comparison. This indicates that simply presenting a reference image does not guarantee that the model identifies the task-relevant counterfactual change.

\paragraph{Paired-image deltas.}
Table~\ref{tab:paired_deltas_by_domain} reports paired-image deltas relative to the counterfactual-only raw-image baseline. Positive $\Delta$Acc indicates improved visual correctness, while negative $\Delta$PFER indicates reduced prior-following. We also report $\Delta$Other because paired inputs may reduce biased answers by increasing uncertainty rather than by improving visual grounding.

\begin{table*}[t]
\centering
\scriptsize
\setlength{\tabcolsep}{2.9pt}
\begin{tabular*}{\textwidth}{@{\extracolsep{\fill}}lccccc@{}}
\toprule
\textbf{Model} & 
\textbf{Existence} & 
\textbf{Counting} & 
\textbf{Fashion} & 
\textbf{Industry} & 
\textbf{Medical} \\
\midrule
GPT-5 
& +7.3/-10.8/+3.5 
& +11.6/-11.8/+0.3 
& +10.5/-9.7/-0.8 
& +1.2/-1.4/+0.1 
& +8.4/-7.5/-0.9 \\

Claude Sonnet 4 
& -8.1/-1.3/+9.4 
& +9.0/-5.6/-3.4 
& -0.4/+1.5/-1.2 
& +8.4/-8.1/-0.3 
& -5.3/+5.2/+0.1 \\

Qwen3-VL-8B-Instruct 
& -11.2/+5.8/+5.4 
& -0.5/-23.6/+24.1 
& +3.9/-2.1/-1.8 
& +21.3/-21.0/-0.3 
& +37.2/-52.4/+15.2 \\

Qwen3-VL-32B-Instruct 
& -16.9/+8.3/+8.5 
& +22.9/-40.8/+17.8 
& +7.9/-7.7/-0.1 
& +13.7/-13.5/-0.2 
& +19.9/-19.6/-0.3 \\

InternVL3.5-8B 
& -29.8/+22.7/+7.1 
& +27.6/-29.9/+2.3 
& +7.1/-5.7/-1.4 
& +9.7/-9.3/-0.4 
& +7.2/+5.7/-12.9 \\

InternVL3.5-38B 
& -26.7/+23.1/+3.5 
& +28.5/-27.3/-1.2 
& +12.5/-11.4/-1.1 
& +5.1/-4.8/-0.3 
& +0.7/-0.7/+0.0 \\

Gemma-3-12B-it 
& -19.2/+17.3/+1.9 
& -6.9/-9.2/+16.1 
& +14.8/-8.0/-6.8 
& +1.6/+0.3/-1.9 
& -30.4/+34.1/-3.7 \\

Gemma-3-27B-it 
& -34.8/+24.6/+10.2 
& +4.4/-14.5/+10.0 
& +22.7/-19.9/-2.7 
& +39.8/-38.8/-1.0 
& +24.8/+1.6/-26.4 \\
\bottomrule
\end{tabular*}
\caption{
\textbf{Paired-image deltas relative to counterfactual-only inputs.} Each cell reports $\Delta$Acc/$\Delta$PFER/$\Delta$Other, in percentage points. Positive $\Delta$Acc and negative $\Delta$PFER indicate improved visual grounding through direct comparison.
}
\label{tab:paired_deltas_by_domain}
\end{table*}

The delta table highlights two patterns. First, paired images often improve performance when the counterfactual change can be directly compared across images, such as counts, visual layouts, or local modality appearance. Second, the paired setting can also hurt performance, especially in object-existence examples, where the presence of the canonical original may increase the salience of the normal object prior.

\paragraph{Model-level summary.}
Table~\ref{tab:paired_model_macro} reports model-level macro-averaged paired-image deltas across domains. Seven of the eight evaluated models improve in accuracy, but the gains vary substantially. Gemma-3-12B-it is the clearest negative case, showing reduced accuracy and increased PFER under paired inputs.

\begin{table}[t]
\centering
\small
\setlength{\tabcolsep}{5pt}
\begin{tabular*}{\linewidth}{@{\extracolsep{\fill}}lrrr@{}}
\toprule
\textbf{Model} &
\textbf{$\Delta$Acc} &
\textbf{$\Delta$PFER} &
\textbf{$\Delta$Other} \\
\midrule
GPT-5 & +7.8 & -8.2 & +0.4 \\
Claude Sonnet 4 & +0.7 & -1.7 & +0.9 \\
Qwen3-VL-8B-Instruct & +10.1 & -18.7 & +8.5 \\
Qwen3-VL-32B-Instruct & +9.5 & -14.7 & +5.1 \\
InternVL3.5-8B & +4.4 & -3.3 & -1.1 \\
InternVL3.5-38B & +4.0 & -4.2 & +0.2 \\
Gemma-3-12B-it & -8.0 & +6.9 & +1.1 \\
Gemma-3-27B-it & +11.4 & -9.4 & -2.0 \\
\bottomrule
\end{tabular*}
\caption{
\textbf{Model-level macro-averaged paired-image deltas across domains.} Values are percentage points relative to counterfactual-only inputs.
}
\label{tab:paired_model_macro}
\end{table}

The model-level summary shows that paired comparison is usually helpful, but not uniformly so. Some models convert paired visual information into higher accuracy and lower PFER, while others show only small changes or even become more prior-driven. This suggests that providing a reference image helps only when the model can identify and use the task-relevant difference.

\paragraph{Domain-level summary.}
Table~\ref{tab:paired_domain_macro} reports domain-level macro-averaged paired-image deltas over models. Counting, fashion, industry/common objects, and medical modality generally benefit from paired comparison. Object existence is the main negative case, suggesting that showing the original object alongside the counterfactual can make the canonical object prior more salient or increase comparison difficulty.

\begin{table}[t]
\centering
\small
\setlength{\tabcolsep}{5pt}
\begin{tabular*}{\linewidth}{@{\extracolsep{\fill}}lrrr@{}}
\toprule
\textbf{Domain} &
\textbf{$\Delta$Acc} &
\textbf{$\Delta$PFER} &
\textbf{$\Delta$Other} \\
\midrule
Existence & -17.4 & +11.2 & +6.2 \\
Counting & +12.1 & -20.3 & +8.3 \\
Fashion & +9.9 & -7.9 & -2.0 \\
Industry & +12.6 & -12.1 & -0.5 \\
Medical & +7.8 & -4.2 & -3.6 \\
\bottomrule
\end{tabular*}
\caption{
\textbf{Domain-level macro-averaged paired-image deltas over models.} Values are percentage points relative to counterfactual-only inputs.
}
\label{tab:paired_domain_macro}
\end{table}

The domain-level summary shows that paired images are most useful when the target difference can be compared directly, such as a count, local layout, visual design feature, or MRI contrast pattern. In contrast, object-existence examples become harder under paired inputs, suggesting that direct comparison may sometimes distract the model or reinforce the canonical object prior.

\begin{figure*}[t]
\centering
\includegraphics[width=\textwidth]{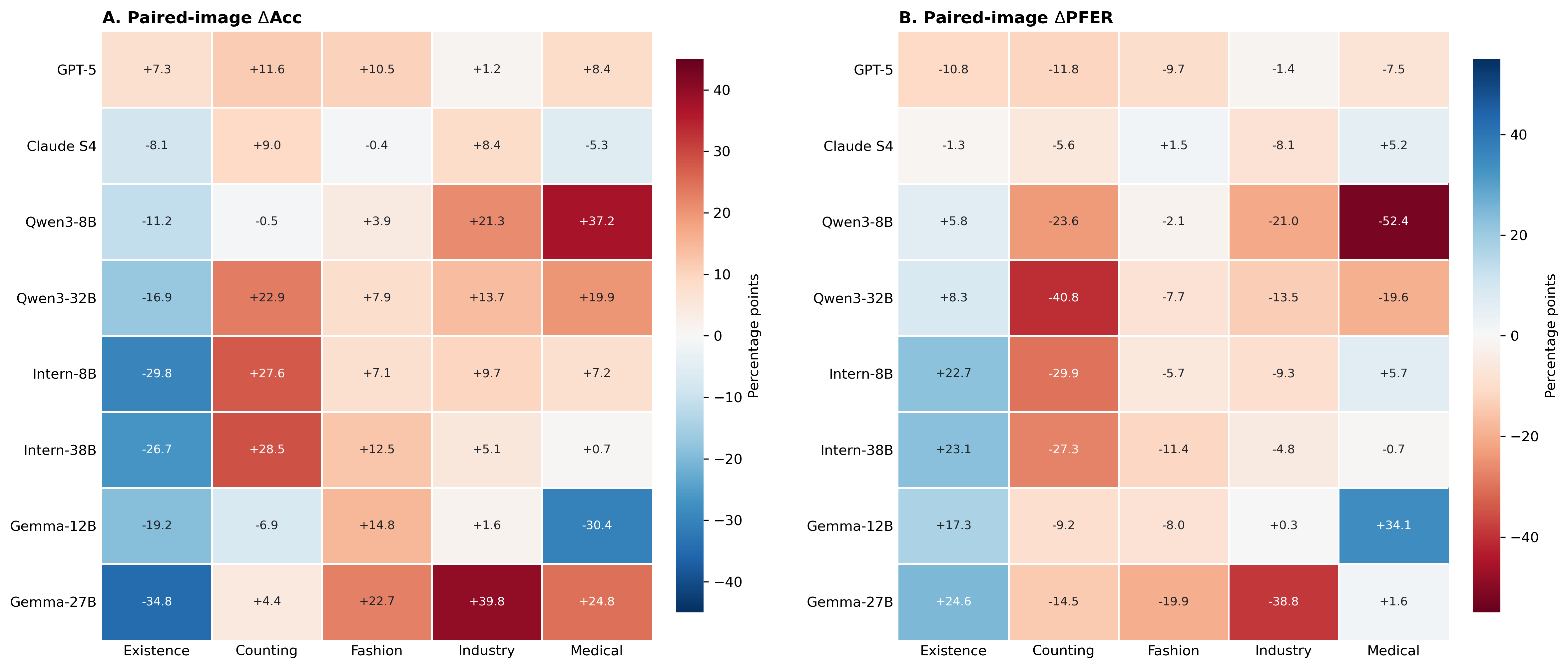}
\caption{
\textbf{Heatmap of paired-image deltas by model and domain.} Panel A reports $\Delta$Acc and Panel B reports $\Delta$PFER relative to counterfactual-only inputs. Positive $\Delta$Acc and negative $\Delta$PFER indicate improved visual grounding through direct comparison.
}
\label{fig:paired_delta_heatmap}
\end{figure*}

\paragraph{Heatmap visualization.}
Figure~\ref{fig:paired_delta_heatmap} visualizes paired-image deltas by model and domain. Panel A shows $\Delta$Acc and Panel B shows $\Delta$PFER. The paired setting is beneficial when accuracy increases while PFER decreases. This pattern appears in several domains, especially counting, fashion, industry/common objects, and parts of the medical modality setting. However, the gains are not uniform: object-existence examples often show lower accuracy and higher PFER under paired inputs.

Overall, paired images provide a useful but limited intervention. They often make counterfactual changes more salient, leading to higher accuracy and lower PFER, but they do not consistently eliminate prior-following behavior. In some cases, paired comparison increases Other Rate or even strengthens prior-following errors, indicating that adding a reference image is not equivalent to reliable visual grounding.

\section{PriVE-Tools Aggregate Results}
\label{app:tool_results}

This appendix reports aggregate PriVE-Tools results used to support RQ3. PriVE-Tools evaluates fixed tool-derived evidence views rather than allowing models to freely call tools. Each tool condition is compared against the matched raw counterfactual baseline for the same model and domain. We report $\Delta$Acc, $\Delta$PFER, and $\Delta$Other. Positive $\Delta$Acc and negative $\Delta$PFER indicate improved grounding, while changes in $\Delta$Other help distinguish genuine correction from increased ambiguity or uncertainty.

For the main cross-domain tool analysis, we focus on the two common evidence conditions available across all five domains: Crop and Zoom panel. This avoids comparing tools with different domain coverage. Table~\ref{tab:aggregate_common_tool_deltas} reports aggregate deltas for all models, closed-source models, and open-source models. Table~\ref{tab:common_tool_deltas_by_model} provides the corresponding model-level breakdown and supports the Top Common Tool column in Table~\ref{tab:main_results}.

\begin{table*}[t]
\centering
\scriptsize
\setlength{\tabcolsep}{4pt}
\begin{tabular*}{\textwidth}{@{\extracolsep{\fill}}llccc@{}}
\toprule
\textbf{Tool} & 
\textbf{Coverage} & 
\textbf{All models} & 
\textbf{Closed-source} & 
\textbf{Open-source} \\
\midrule
Crop 
& 5 domains / 8 models 
& -0.9 / +0.2 / +0.7 
& +7.3 / -6.9 / -0.4 
& -3.6 / +2.6 / +1.0 \\

Zoom panel 
& 5 domains / 8 models 
& -0.2 / -1.1 / +1.3 
& +8.0 / -7.5 / -0.4 
& -3.0 / +1.1 / +1.9 \\
\bottomrule
\end{tabular*}
\caption{
\textbf{Aggregate PriVE-Tools deltas for common evidence conditions available across all five domains.} Each metric cell reports $\Delta$Acc / $\Delta$PFER / $\Delta$Other, in percentage points, relative to the matched raw counterfactual baseline. A useful evidence condition should increase accuracy and decrease PFER without substantially increasing Other.
}
\label{tab:aggregate_common_tool_deltas}
\end{table*}

Table~\ref{tab:aggregate_common_tool_deltas} shows that common evidence views do not uniformly improve grounding. Across all evaluated models, Crop and Zoom panel have small or negative effects on accuracy. However, this aggregate pattern masks a strong group-level split: the same evidence views improve accuracy and reduce PFER for closed-source models, but are weak or negative on average for open-source models.

\paragraph{Model-level common tool deltas.}
Table~\ref{tab:common_tool_deltas_by_model} reports common tool-condition deltas for each model. These values are macro-averaged across the five domains and computed relative to each model's matched raw counterfactual baseline. The table supports the Top Common Tool column in Table~\ref{tab:main_results}, where the best common tool is selected by the largest macro-averaged accuracy gain among Crop and Zoom panel.

\begin{table*}[t]
\centering
\scriptsize
\setlength{\tabcolsep}{5pt}
\begin{tabular*}{\textwidth}{@{\extracolsep{\fill}}lcc@{}}
\toprule
\textbf{Model} & 
\textbf{Crop} & 
\textbf{Zoom panel} \\
\midrule
\multicolumn{3}{l}{\textit{Closed-source VLMs}} \\
GPT-5 
& +4.5 / -4.3 / -0.2 
& +4.2 / -4.4 / +0.2 \\

Claude Sonnet 4 
& +10.1 / -9.5 / -0.6 
& +11.8 / -10.6 / -1.1 \\

\midrule
\multicolumn{3}{l}{\textit{Open-source VLMs}} \\
Qwen3-VL-8B-Instruct 
& -2.5 / +3.4 / -0.9 
& -1.4 / +1.2 / +0.3 \\

Qwen3-VL-32B-Instruct 
& -2.2 / +2.9 / -0.7 
& -2.8 / +3.3 / -0.5 \\

InternVL3.5-8B 
& -2.4 / -0.8 / +3.2 
& +0.5 / -5.8 / +5.2 \\

InternVL3.5-38B 
& -5.7 / -2.8 / +8.5 
& -8.3 / -1.8 / +10.1 \\

Gemma-3-12B-it 
& -7.2 / +9.2 / -2.0 
& -5.2 / +6.6 / -1.5 \\

Gemma-3-27B-it 
& -1.6 / +3.6 / -2.0 
& -0.7 / +2.8 / -2.1 \\
\bottomrule
\end{tabular*}
\caption{
\textbf{Common tool-condition deltas by model.} Each cell reports $\Delta$Acc / $\Delta$PFER / $\Delta$Other, in percentage points, macro-averaged across the five domains and computed relative to the matched raw counterfactual baseline.
}
\label{tab:common_tool_deltas_by_model}
\end{table*}

Table~\ref{tab:common_tool_deltas_by_model} shows that the same evidence condition can have different effects across models. GPT-5 and Claude Sonnet 4 both benefit from common tool views, especially Claude Sonnet 4 under Zoom panel evidence. In contrast, several open-source models show negative $\Delta$Acc even under their better common tool condition. Some models reduce PFER while increasing Other Rate, indicating that tool evidence may shift responses from prior-following to uncertainty rather than to correct visual grounding.

\begin{figure*}[t]
\centering
\includegraphics[width=\linewidth]{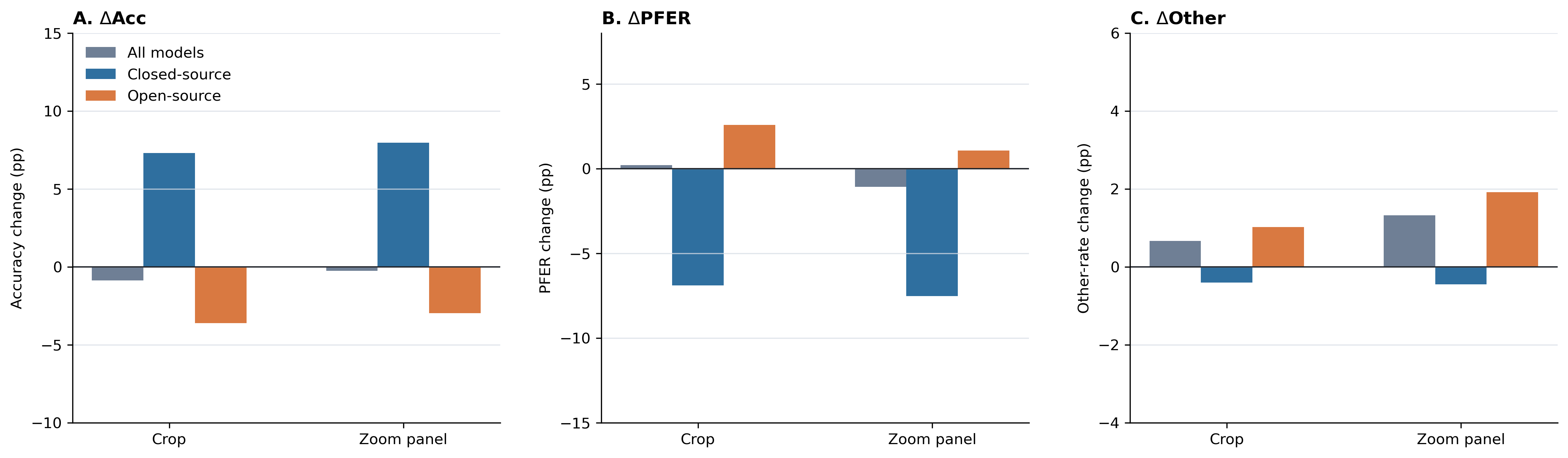}
\caption{
\textbf{Macro-averaged effects of common PriVE-Tools evidence conditions across models and domains.} Bars report $\Delta$Acc, $\Delta$PFER, and $\Delta$Other relative to matched raw counterfactual baselines. Positive $\Delta$Acc and negative $\Delta$PFER indicate improved grounding.
}
\label{fig:common_tool_delta_bars}
\end{figure*}

\paragraph{Aggregate visualization.}
Figure~\ref{fig:common_tool_delta_bars} visualizes the aggregate effects of the common tool conditions. The figure separates all models, closed-source models, and open-source models. Positive $\Delta$Acc and negative $\Delta$PFER indicate improved grounding.

Figure~\ref{fig:common_tool_delta_bars} reinforces the main RQ3 finding: common tool-derived evidence is beneficial for closed-source models in our evaluation, but weak or negative on average for open-source models. This suggests that providing localized or magnified evidence is not sufficient by itself; the model must also be able to integrate that evidence into the answer.

\paragraph{Partially covered and domain-specific tools.}
Some PriVE-Tools evidence conditions are domain-specific or only partially covered in the current experiments. Counting includes outline overlays, medical modality includes contour overlays, and bounding-box overlays are incomplete for Fashion and Industry because the open-source bbox runs were not completed. We therefore report these results separately in Table~\ref{tab:domain_specific_tool_deltas} rather than using them for the main cross-domain Top Common Tool selection.

\begin{table*}[t]
\centering
\scriptsize
\setlength{\tabcolsep}{3.5pt}
\begin{tabular*}{\textwidth}{@{\extracolsep{\fill}}llp{3.0cm}ccc@{}}
\toprule
\textbf{Tool} & 
\textbf{Coverage} & 
\textbf{Domains} & 
\textbf{All models} & 
\textbf{Closed-source} & 
\textbf{Open-source} \\
\midrule
BBox 
& 28 model-domain pairs 
& Counting, Existence, Fashion, Industry, Medical 
& -2.0 / -0.1 / +2.1 
& +5.5 / -5.6 / +0.2 
& -6.1 / +2.9 / +3.2 \\

Outline 
& 8 model-domain pairs 
& Counting 
& -4.7 / +2.7 / +1.9 
& +0.7 / -2.7 / +2.1 
& -6.5 / +4.6 / +1.9 \\

Contour 
& 8 model-domain pairs 
& Medical 
& +3.0 / -7.3 / +4.3 
& +14.3 / -13.1 / -1.2 
& -0.8 / -5.4 / +6.2 \\
\bottomrule
\end{tabular*}
\caption{
\textbf{Domain-specific or partially covered PriVE-Tools deltas.} Outline is counting-specific and Contour is medical-specific. Each metric cell reports $\Delta$Acc / $\Delta$PFER / $\Delta$Other relative to the matched raw counterfactual baseline.
}
\label{tab:domain_specific_tool_deltas}
\end{table*}

Table~\ref{tab:domain_specific_tool_deltas} shows that some domain-specific evidence views can be useful, especially for closed-source models. For example, Contour evidence in the medical domain substantially improves the closed-source aggregate, while its open-source effect is weaker and accompanied by increased Other Rate. Because these tools do not share uniform coverage across all domains and models, we treat them as supplementary evidence rather than as the basis for the main cross-domain tool comparison.

Overall, these aggregate results support a cautious interpretation of PriVE-Tools. Tool-derived evidence can improve grounding, especially for stronger closed-source models and for some domain-specific evidence views. However, tools do not consistently convert prior-following errors into correct answers. In several open-source settings, tool views reduce accuracy, increase PFER, or increase Other Rate. This motivates the domain- and model-specific analysis in RQ4.

\section{Domain- and Model-Specific Tool Analysis}
\label{app:tool_analysis}

This appendix provides the domain- and model-level PriVE-Tools analysis used to support RQ4. While Appendix~\ref{app:tool_results} reports aggregate tool effects, this section examines where those effects come from. We focus on displayed evidence conditions that are directly interpretable across the benchmark: bounding-box overlays, object crops, and zoom panels. All deltas are computed relative to the matched raw counterfactual baseline for the same model and domain.

Because the Fashion and Industry bbox runs were not completed for all open-source models, those cells are left blank in the domain-level analysis rather than treated as complete evidence. For model-level bbox summaries, we macro-average only over domains with complete bbox coverage across all eight main models: object existence, counting, and medical modality. This keeps model-level comparisons consistent.

\paragraph{Best displayed tool by domain.}
Table~\ref{tab:best_displayed_tool_by_domain} reports the best displayed tool by domain, selected by macro-averaged $\Delta$Acc among available displayed evidence views. The corresponding $\Delta$PFER and $\Delta$Other values indicate whether the gain reflects improved grounding, reduced prior-following, or increased uncertainty.

\begin{table*}[t]
\centering
\scriptsize
\setlength{\tabcolsep}{4pt}
\begin{tabular*}{\textwidth}{@{\extracolsep{\fill}}llcccp{6.3cm}@{}}
\toprule
\textbf{Domain} & 
\textbf{Best displayed tool} & 
\textbf{$\Delta$Acc} & 
\textbf{$\Delta$PFER} & 
\textbf{$\Delta$Other} & 
\textbf{Interpretation} \\
\midrule
Existence 
& Crop 
& -4.3 
& +5.3 
& -1.0 
& Common displayed tools do not improve object-existence judgments on average; localized views can still leave models anchored to category priors. \\

Counting 
& Zoom panel 
& -3.1 
& +1.0 
& +2.1 
& Magnified evidence does not by itself solve enumeration, and can shift some prior-following errors into uncertainty. \\

Fashion 
& Zoom panel 
& -2.5 
& +3.0 
& -0.5 
& Fine-grained logo and element edits remain prior-sensitive on average, even when local visual evidence is provided. \\

Industry 
& Crop 
& +4.4 
& -4.1 
& -0.2 
& Local crops help most clearly for common-object attribute changes, where the manipulated object can be inspected directly. \\

Medical 
& Zoom panel 
& +5.3 
& -10.5 
& +5.2 
& Zoom panels reduce modality-prior errors, but the increased Other rate shows that some gains reflect uncertainty around local contrast. \\
\bottomrule
\end{tabular*}
\caption{
\textbf{Best-performing displayed tool condition by domain.} The best tool is selected by macro-averaged $\Delta$Acc among available displayed tools, with corresponding changes in PFER and Other reported to indicate whether improvements reflect better grounding or increased uncertainty. Deltas are relative to matched raw counterfactual baselines. Fashion and Industry BBox cells are excluded because open-source bbox runs were not completed.
}
\label{tab:best_displayed_tool_by_domain}
\end{table*}

Table~\ref{tab:best_displayed_tool_by_domain} shows that no evidence view dominates across domains. Local crops are most useful for industry/common-object attributes, where the manipulated property is localized and visually inspectable. Zoom panels help medical modality consistency by exposing local tumor-region contrast, but the increased Other Rate suggests that some models become uncertain rather than fully grounded. Object existence, counting, and fashion remain challenging even when the relevant region is localized or magnified.

\begin{figure*}[t]
\centering
\includegraphics[width=\linewidth]{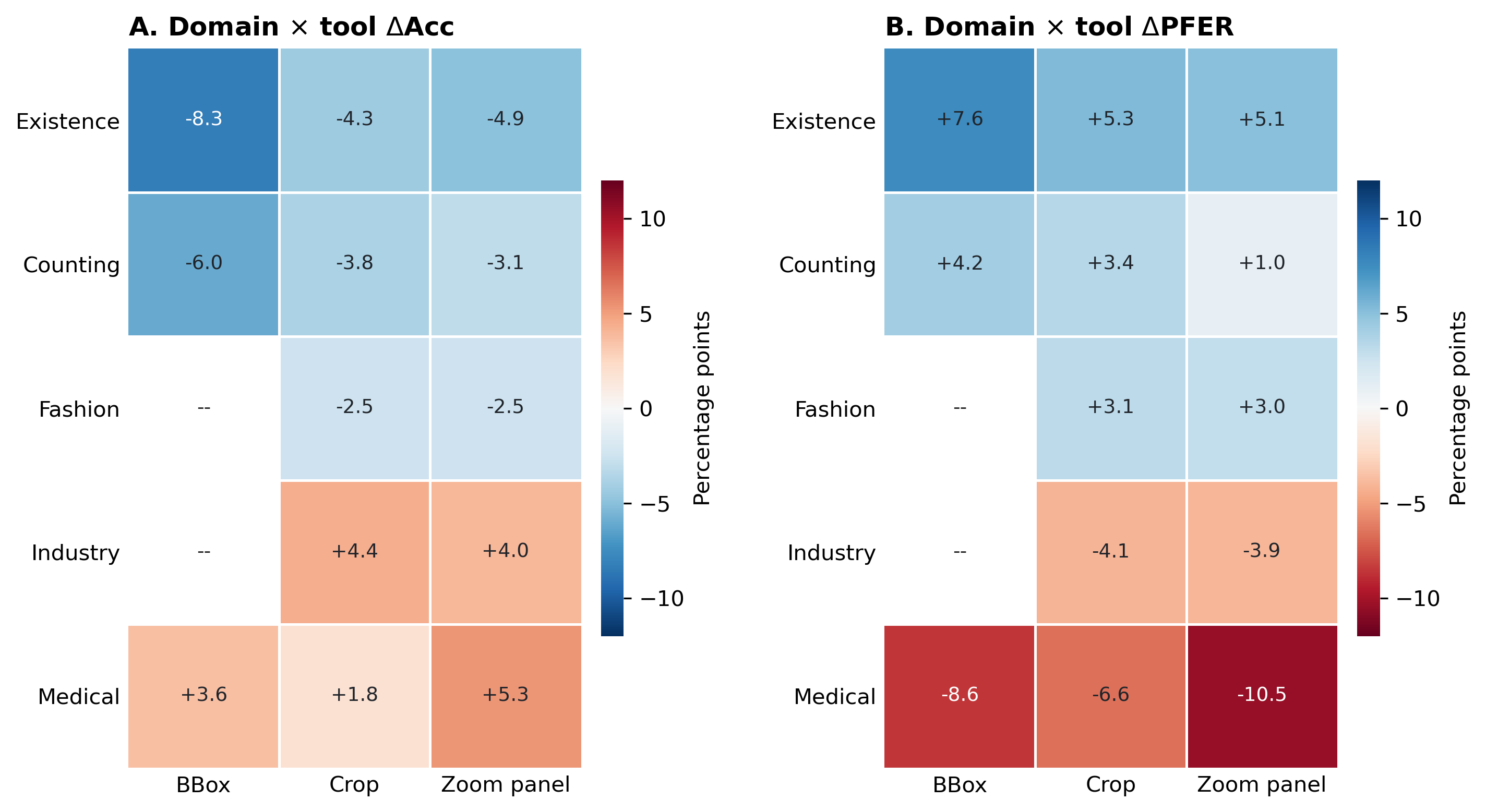}
\caption{
\textbf{Domain-by-tool heatmaps for displayed PriVE-Tools evidence conditions.} Panel A reports $\Delta$Acc and Panel B reports $\Delta$PFER relative to matched raw counterfactual baselines. Fashion and Industry BBox cells are left blank because open-source bbox runs were not completed.
}
\label{fig:domain_tool_heatmap}
\end{figure*}

\paragraph{Domain-level heatmaps.}
Figure~\ref{fig:domain_tool_heatmap} visualizes domain-level tool effects. Panel A reports $\Delta$Acc and Panel B reports $\Delta$PFER relative to matched raw counterfactual baselines. The strongest positive displayed-tool effects appear in Industry and Medical. By contrast, object existence, counting, and fashion show weak or negative gains under the displayed evidence views, indicating that localized evidence alone does not solve absence reasoning, enumeration, or fine-grained prior-resistant recognition.

\begin{figure*}[t]
\centering
\includegraphics[width=\linewidth]{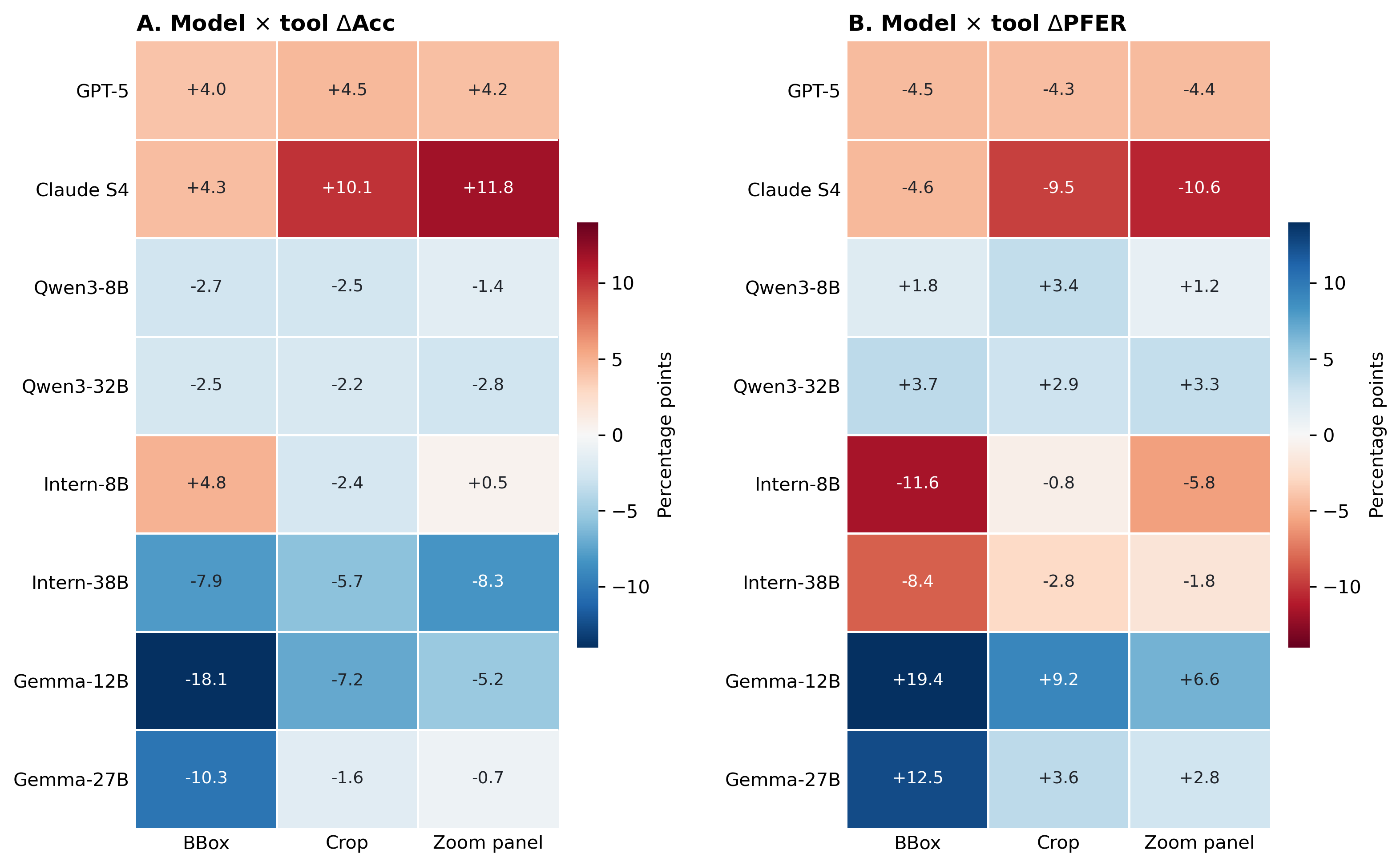}
\caption{
\textbf{Model-by-tool heatmaps for displayed PriVE-Tools evidence conditions.} Panel A reports $\Delta$Acc and Panel B reports $\Delta$PFER. BBox values are macro-averaged over domains with complete bbox coverage across all models.
}
\label{fig:model_tool_heatmap}
\end{figure*}

\paragraph{Model-level heatmaps.}
Figure~\ref{fig:model_tool_heatmap} shows the corresponding model-level pattern. Closed-source models benefit more consistently from localized evidence: GPT-5 and Claude Sonnet 4 show positive accuracy gains and reduced PFER under Crop and Zoom panel. Several open-source models, however, show negative accuracy changes, increased PFER, or increased Other Rate under the same evidence conditions. This indicates that tool usefulness depends not only on whether relevant visual evidence is visible, but also on whether the model can use that evidence against learned priors.

\paragraph{Summary.}
Together, these results show that PriVE-Tools effects are both domain-specific and model-specific. Local crops are most useful for common-object attribute changes, zoom panels help expose medical modality conflicts but may increase uncertainty, and bounding boxes do not uniformly help object-existence or counting tasks. The same evidence view can therefore help one model or domain while degrading another. This supports the main conclusion that agentic visual evidence is not a universal remedy: it exposes relevant evidence, but VLMs still need the ability to interpret and use that evidence correctly.

\end{document}